%% file: main.tex
\documentclass{article}
\usepackage[preprint]{log_2022}
\usepackage{booktabs}            
\usepackage{multirow}            
\usepackage{amsfonts}            
\usepackage{graphicx}            
\usepackage{duckuments}          
\usepackage{array}
\usepackage{wrapfig}

\usepackage[numbers,compress,sort]{natbib}
\definecolor{Gray}{gray}{0.95}

\usepackage{amssymb}
\usepackage{amsthm}
\usepackage[linesnumbered,ruled,vlined,noend]{algorithm2e}
\usepackage{subcaption}

\usepackage{macros}
\usepackage[table]{colortbl}

\title[Graph Coarsening via Convolution Matching \\ for Scalable Graph Neural Network Training]{Graph Coarsening via Convolution Matching \\ for Scalable Graph Neural Network Training}

\author[C. Dickens et al.]{%
Charles Dickens\\
\institute{University of California, Santa Cruz}\\
\email{cadicken@ucsc.edu}\\
\And
Eddie Huang\\
\institute{Amazon}\\
\email{ewhuang@amazon.com}\\
\And
Aishwarya Reganti\\
\institute{Amazon}\\
\email{areganti@amazon.com}\\
\And
Jiong Zhu\\
\institute{University of Michigan}\\
\email{jiongzhu@umich.edu}\\
\And
Karthik Subbian\\
\institute{Amazon}\\
\email{ksubbian@amazon.com}\\
\And
Danai Koutra\\
\institute{Amazon}\\
\email{dkoutra@amazon.com}
}

\begin{document}

\maketitle

\begin{abstract}
  \input{sections/abstract}
\end{abstract}

\input{sections/introduction}
\input{sections/related_works}
\input{sections/background}
\input{sections/methodology}

\input{sections/evaluation}
\input{sections/conclusion}

\bibliographystyle{unsrtnat}
\bibliography{main}

\newpage 
\appendix

\input{sections/appendix}

\end{document}

%% file: sections/abstract.tex
Graph summarization as a preprocessing step is an effective and complementary technique for scalable graph neural network (GNN) training.
In this work, we propose the \longname{} (\shortname{}) algorithm and a highly scalable variant, A-\shortname{}, for creating summarized graphs that preserve the output of 
graph convolution.
We evaluate \shortname{} on six real-world link prediction and node classification graph datasets, and show it is efficient and preserves prediction performance while significantly reducing the graph size.
Notably, \shortname{} achieves up to 95\% of the prediction performance of GNNs on node classification while trained on graphs summarized down to 1\% the size of the original graph.
Furthermore, on link prediction tasks, \shortname{} consistently outperforms all baselines, achieving up to a $2 \times$ improvement.

%% file: sections/introduction.tex
\section{Introduction}

Graph neural networks (GNNs) have achieved state-of-the-art performance on various tasks ranging from recommendation to predicting drug interactions~\cite{hu2020ogb}.
However, a drawback of a GNN's modeling capacity is a computationally expensive inference process with a complexity that scales with the size of the graph.
Modern techniques for scaling the training of deep models, such as leveraging the parallel structure of GPUs for processing large blocks of data, have been successfully adopted by the GNN community~\cite{wang2019dgl,Fey2019PyG}.
However, graph datasets encountered in real-world applications are on the order of tens of billions of edges~\cite{zheng2021distributed} and quickly exceed the costly and limited memory capacity of today's GPUs.
Techniques for partitioning and distributing the training graph across computational resources and integrating graph sampling into the training pipeline have been proposed~\cite{zheng2020distdgl,jiang2021communication,ramezani2021learn,zhu2023simplifying}.
Nonetheless, training GNNs is still a highly expensive process, which limits applicability and prohibits large-scale architecture searches. 
Furthermore, distributed training and sampling techniques introduce their own difficulties.
For example, distributed training faces communication overhead across machines~\cite{gandhi2021p3}, while sampling techniques bring additional hyperparameters that affect model performance~\cite{ramezani2021learn,zhu2023simplifying}.

A promising new direction of scalable GNN training is to perform graph summarization, i.e., create a smaller graph with fewer nodes and edges, as a preprocessing step.
These methods either sample nodes and edges from the original training graph \citep{tsang:jmlr05, har-peled:acg05, welling:icml09, sener:iclr18, mirzasoleiman:icml20}, coarsen the original graph by clustering nodes into supernodes \citep{huang:kdd21}, or create synthetic graph connections and node features \citep{zhao:iclr21, jin:kdd22}.
To be applicable for scalable GNN training, the graph summarization process should be faster than fitting a GNN on the original graph. 
Additionally, the summarized graph should share properties with the original graph such that a GNN can be fit for various downstream tasks with good performance.
Existing approaches to graph summarization typically fail to satisfy at least one of the mentioned desirable properties (Table~\ref{tab:baseline_comparison}).

In this work, we introduce \longname{} (\shortname{}), a highly scalable graph coarsening algorithm. 
\shortname{} iteratively merges supernodes that minimize a cost quantifying the change in the output of graph convolution.
Notably, \shortname{} merges supernodes that are structurally similar, allowing the algorithm to identify and summarize redundant nodes that are distant or even disconnected in the original graph. 
Our primary contributions in this work are:
\begin{itemize}
\item \textbf{New Approach}: We introduce \shortname{}, a coarsening algorithm that preserves the output of graph convolutions.
\item \textbf{Highly-scalable Variant}: We propose a principled approximation to computing costs in the \shortname{} algorithm, A-\shortname{}, which allows it to scale to large graphs.
\item \textbf{Extensive Empirical Analysis}: We perform an extensive empirical analysis demonstrating our method's ability to summarize large-scale graphs and preserve GNN prediction performance. 
On link prediction tasks, \shortname{} achieves up to a $2 \times$ prediction performance improvement over the best baseline.
In node classification, it achieves up to 95\% of the prediction performance of a GNN on a graph that is 1\% the size of the original.
\end{itemize}

%% file: sections/related_works.tex
\section{Related Work}
\label{sec:related_works}

\input{sections/tables/baseline_comparison}

We give a qualitative comparison of our \shortname{} variants to the most relevant work in Table~\ref{tab:baseline_comparison}.

\noindent \textbf{Coreset Selection}.
Coreset methods aim to find a subset of training examples such that a model trained on the subset will perform similarly to a model trained on the complete dataset \citep{tsang:jmlr05, har-peled:acg05, welling:icml09, sener:iclr18, mirzasoleiman:icml20}.
\emph{Herding}, proposed by \citenoun{welling:icml09}~\cite{welling:icml09} is a coreset technique in which training examples are first mapped to an embedding space and then clustered by class.
Examples closest to the cluster center in the embedding space are selected.
The \emph{KCenter} algorithm~\cite{sener:iclr18} 
similarly embeds training data and then incrementally selects examples with the largest minimum distance to the growing cluster subset.

\noindent \textbf{Graph Condensation}.
Distillation and condensation techniques search for a small synthetic dataset such that model parameters fit on the synthetic dataset are approximate minimizers of the training objective on the original dataset \citep{wang:arxiv18, bohdal:arxiv20, sucholutsky:arxiv20, nguyen:iclr21, zhao:iclr21, zhao:icml21, jin:iclr22, jin:kdd22}.
Recently, \citenoun{jin:iclr22}~\cite{jin:iclr22} extended the dataset condensation via gradient matching scheme proposed by \citenoun{zhao:iclr21}~\cite{zhao:iclr21} with the \emph{GCond} algorithm, which synthesizes graph data for training GNNs, and later with \emph{DosCond}, which performs one-step gradient matching to find the synthesized graph~\cite{jin:kdd22}.
Alternatively, \citenoun{liu:arxiv22}~\cite{liu:arxiv22} propose a condensation method for creating a synthetic graph that aims to match statistics of the receptive field of the original graph nodes. 
\shortname{} is not a graph condensation method as it does not generate a fully synthetic graph by matching graph statistics or learning loss gradients of training nodes computed with the original graph structure.
\shortname{} does use a structural embedding to create node pairs initially; however, the embedding does not need to be learned.

\noindent \textbf{Graph Coarsening}.
Coarsening is a graph summarization \citep{liu:compserv19} technique in which nodes and/or edges from an original graph are merged to form a supergraph.
Graph coarsening methods are widely applied and studied for problems ranging from influence analysis \citep{purohit:kdd14}, visualization \citep{harel:jgaa02, walshaw:jgaa06, shneiderman:sigchi13}, combinatorial optimization \citep{moitra:sfcs09, englert:siamcomp14}, and, recently, scaling graph embeddings \citep{liang:icwsm21, akbas:ieeebd19, fahrbach:icml20, zhang:ieeekde20, deng:iclr20, huang:kdd21, kumar:icml23}. 
Moreover, coarsening methods typically have the practically advantageous property of producing multi-level summaries, i.e., producing summaries at multiple level of granularity.
\citenoun{huang:kdd21}~\cite{huang:kdd21} specifically proposed coarsening to overcome scalability issues of GNN training.
The authors coarsen the graph used for training the GNN, with algorithms by \citenoun{loukas:jmlr19}~\cite{loukas:jmlr19}.
\shortname{} is a graph coarsening algorithm that aims to preserve the graph convolution operations that are fundamental to spectral-based GNNs.
\shortname{} differs from existing coarsening approaches for scalable GNN training in that a structural embedding is used to initially pair nodes to consider for merging rather than pair nodes based on their proximity in the original graph.

%% file: sections/tables/baseline_comparison.tex
\begin{table}[t]
    \centering
    \caption{Qualitative comparison of graph summarization methods.
    `Summary': the type of summarized graph produced;
    `NC/LP': summarized graph can be used to train a model for node classification or link prediction, resp.;
    `No GNN on full graph': does not require fitting a GNN on the full graph;
    `Multi-level': multiple levels of summarization produced;
    `Merge Strategy': if applicable, the strategy for selecting nodes to merge.
    }
    \scalebox{0.6}{
    \begin{tabular}{lll@{\hskip 0.5cm}|cccccc}
    \toprule
        \textbf{Core Technique} & \textbf{Summary} & \textbf{Method} & \textbf{NC} & \textbf{LP} & \shortstack[l]{\textbf{No GNN on} \\ \textbf{Full Graph}} & \textbf{Multi-Level} & \textbf{Merge Strategy} \\
    \midrule
        \multirow{3}{*}{Coreset} & \multirow{3}{*}{\shortstack[l]{Sampled\\graph}}
        & RS & \checkmark & \checkmark & \checkmark & \checkmark & \multirow{3}{*}{-} \\
        & & KCenter~\cite{sener:iclr18} & \checkmark & $ \times $ & $ \times $ & \checkmark &  \\
        & & Herding~\cite{welling:icml09} & \checkmark & $ \times $ & $ \times $ & \checkmark & \\
    \midrule
        \multirow{2}{*}{Condensation} & \multirow{2}{*}{\shortstack[l]{Synthetic\\ graph}}
        & GCond~\cite{jin:iclr22} & \checkmark & \checkmark & $ \times $ & $ \times $ & \multirow{2}{*}{-} \\
        & & DosCond~\cite{jin:kdd22} & \checkmark & \checkmark & $ \times $ & $ \times $ &  \\
    \midrule
        \multirow{3}{*}{Coarsening} & \multirow{3}{*}{Supergraph}
        & VN~\cite{loukas:jmlr19} & \checkmark & \checkmark & \checkmark & \checkmark & Neighbors \\
        & & \shortname{} (ours) & \checkmark & \checkmark & \checkmark & \checkmark & Structurally similar \\
        & & A-\shortname{} (ours) & \checkmark & \checkmark & \checkmark & \checkmark & Structurally similar \\
    \bottomrule
    \end{tabular}
    }
    \label{tab:baseline_comparison}
    \vspace{-0.4cm}
\end{table}

%% file: sections/background.tex
\section{Preliminaries}
We start with key notations and the necessary background for describing our proposed approach.

\par{\textbf{Graph Notations}.} Let $G = (\mathcal{V}, \mathcal{E})$ denote a graph with a node attribute matrix $\mathbf{X} \in \mathbb{R}^{n \times d}$, where $n = \vert \mathcal{V} \vert$ and $d > 0$.
Moreover, let $\mathbf{A} \in \{0, 1\}^{n \times n}$ be the adjacency matrix corresponding to the graph $G$, and $\mathbf{D}$ be the diagonal degree matrix. 

\par{\textbf{Graph Coarsening}.}
A \emph{coarse graph} is defined from a partitioning of the nodes into $n' \leq n$ clusters: $\mathcal{P} = \{\mathcal{C}_1, \mathcal{C}_2, \cdots, \mathcal{C}_{n'} \}$.
Each partition, $\mathcal{C}_i \in \mathcal{P}$, is referred to as a \emph{supernode}.
The partitioning is represented by a partition matrix $\mathbf{P} \in \left\{ \mathbf{P}' \in \{0, 1\}^{n \times n'} \Big \vert \, \sum_{j} \mathbf{P}'_{i, j} = 1, \, \forall i \right\} \triangleq \mathbb{P}(n, n')$, where entry  $\mathbf{P}_{i, j} = 1$ if and only if $v_{i} \in \mathcal{C}_j$.
Given the partition matrix, the coarse graph $G' = (\mathcal{V}', \mathcal{E}')$ is constructed with an adjacency matrix $\mathbf{A}' \triangleq \mathbf{P}^T \mathbf{A} \mathbf{P}$ and degree matrix $\mathbf{D}' \triangleq \mathbf{P}^T \mathbf{D} \mathbf{P}$.
We define the supernode size matrix of the coarse graph as  $\mathbf{C} \triangleq \textrm{diag}([\vert \mathcal{C}_{1} \vert, \vert \mathcal{C}_{2} \vert, \cdots, \vert \mathcal{C}_{n'} \vert])$.
Then, the coarse node attribute matrix is given as $\mathbf{X}' \triangleq \mathbf{C}^{-1} \mathbf{P}^T \mathbf{X}$.

\par{\textbf{Spectral Graph Convolutions}.}
Spectral-based GNNs are a prominent class of models rooted in graph Fourier analysis \citep{bruna:iclr14, defferrard:nips16, kipf:iclr17, levie:sp18, wu:icml19}.
These methods generally assume graphs to be undirected and rely on the graph Laplacian: $\mathbf{\Delta} \triangleq \mathbf{D} - \mathbf{A}$, and its eigendecomposition: $\mathbf{\Delta} = \mathbf{U} \mathbf{\Lambda} \mathbf{U}^T$, where $\mathbf{U} \in \mathbb{R}^{n \times n}$ is an orthonormal matrix comprising the eigenvectors of $\mathbf{\Delta}$, and $\Lambda = \textrm{diag}(\lambda_1, \cdots, \lambda_n)$ is the diagonal matrix of eigenvalues. 
The \emph{graph Fourier transform} of a signal $\mathbf{x} \in \mathbb{R}^{n}$ over the graph $G$ is defined as $\mathcal{F}_{G}(\mathbf{x}) \triangleq \mathbf{U}^T \mathbf{x}$.
As $\mathbf{U}$ is an orthonormal matrix, the inverse graph Fourier transform is thus $\mathcal{F}_{G}^{-1}(\mathbf{x}) \triangleq \mathbf{U} \mathbf{x}$.
The \emph{graph convolution} of a signal $\mathbf{x} \in \mathbb{R}^{n}$ and a signal, or filter, $\mathbf{g} \in \mathbb{R}^{n}$, is the inverse transform of the element-wise product ($\odot$) of the signals in the transformed domain: 
\begin{align}
    \mathbf{g} \star_{G} \mathbf{x} 
    \triangleq \mathcal{F}_{G}^{-1} (\mathcal{F}_{G}(\mathbf{x}) \odot \mathcal{F}_{G}(\mathbf{g}))
    = \mathbf{U} (\mathbf{U}^T \mathbf{x} \odot \mathbf{U}^T \mathbf{g}).
\end{align}
Spectral-based GNNs use this definition to motivate architectures that approximate graph convolutions and parameterize the filter.
We write $\mathbf{g}_{\theta}$ to denote a filter parameterized by a scalar $\theta$.
\citenoun{kipf:iclr17}\cite{kipf:iclr17} make the following principled approximation of graph convolution:
\begin{align}
    \mathbf{g}_{\theta} \star_{G} \mathbf{x} \approx \theta (\tilde{\mathbf{D}}^{-\frac{1}{2}} \tilde{\mathbf{A}} \tilde{\mathbf{D}}^{-\frac{1}{2}}) \mathbf{x},
    \label{eq:gcn_convolution}
\end{align}
where $\tilde{\mathbf{A}} = \mathbf{A} + \mathbf{I}$ (graph with self-loops) and $\tilde{\mathbf{D}}$ is the degree matrix of $\tilde{\mathbf{A}}$.
A $K$-layer graph convolutional network (GCN) is a recursive application of \eqref{eq:gcn_convolution} and a activation function, $\sigma(\cdot)$:
\begin{align}
    \mathbf{H}^{(K)} &\triangleq 
    \begin{cases}
        \mathbf{X} & K = 0 \\
        \sigma((\tilde{\mathbf{D}}^{-\frac{1}{2}} \tilde{\mathbf{A}} \tilde{\mathbf{D}}^{-\frac{1}{2}}) \mathbf{H}^{(K - 1)} \mathbf{\Theta}^{(K)}) & \textrm{o.w.}
    \end{cases}
\end{align}
Note that the graph signal is generalized to an $(n \times f)$ matrix $\mathbf{X}$ and each GCN layer is parameterized by a matrix $\mathbf{\Theta}^{(K)}$.
Furthermore, defining $\tilde{\mathbf{H}}^{(K)} \triangleq (\tilde{\mathbf{D}}^{-\frac{1}{2}} \tilde{\mathbf{A}} \tilde{\mathbf{D}}^{-\frac{1}{2}}) \mathbf{H}^{(k - 1)}$ and $\mathbf{H}^{(0)} \triangleq \mathbf{X}$, we have the equivalent compact form:
$
    \mathbf{H}^{(K)} \triangleq \sigma(\tilde{\mathbf{H}}^{(K)} \mathbf{\Theta}^{(K)}) 
    \label{eq:gcn_layer}
$.

A notable instantiation of the GCN architecture proposed by \citenoun{wu:icml19}\cite{wu:icml19} is the simplified GCN (SGC), which uses the identity operator as the activation.
$K$ recursive applications of SGC layers is equivalent to a single linear operator acting on $(\tilde{\mathbf{D}}^{-\frac{1}{2}} \tilde{\mathbf{A}} \tilde{\mathbf{D}}^{-\frac{1}{2}})^{K} \mathbf{X}$:
\begin{align}
    \mathbf{H}_{SGC}^{(K)} \triangleq (\tilde{\mathbf{D}}^{-\frac{1}{2}} \tilde{\mathbf{A}} \tilde{\mathbf{D}}^{-\frac{1}{2}})^{K} \mathbf{X} \mathbf{\Theta}.
    \label{eq:SGC_network}
\end{align}
This expression 
illustrates the primary benefits of the SGC architecture; 
the result of $(\tilde{\mathbf{D}}^{-\frac{1}{2}} \tilde{\mathbf{A}} \tilde{\mathbf{D}}^{-\frac{1}{2}})^{K} \mathbf{X}$ is cached so future inferences do not require computation of the intermediate representations of nodes.
Moreover, the parameter space reduces to a single matrix $\mathbf{\Theta}$. 

\par{\textbf{Coarse Graph Convolutions}.}
\citenoun{huang:kdd21}\cite{huang:kdd21} propose coarse graph convolution layers.
Setting $\tilde{\mathbf{A}}' \triangleq \mathbf{A}' + \mathbf{C}$ and $\tilde{\mathbf{D}}' \triangleq \mathbf{D}' + \mathbf{C}$, a coarse graph convolution is recursively defined as
\begin{align}
    \mathbf{H}'^{(K)} \triangleq 
    \begin{cases}
        \mathbf{X}' & K = 0 \\
        \sigma((\tilde{\mathbf{D}}'^{-\frac{1}{2}} \tilde{\mathbf{A}}' \tilde{\mathbf{D}}'^{-\frac{1}{2}}) \mathbf{H}'^{(K - 1)} \mathbf{\Theta}^{(K)}) & \textrm{o.w.}
    \end{cases}
\end{align}
Similar to GCN convolutions, defining $\mathbf{H}'^{(0)} \triangleq \mathbf{X}'$ and $\tilde{\mathbf{H}}'^{(K)} \triangleq (\tilde{\mathbf{D}}'^{-\frac{1}{2}} \tilde{\mathbf{A}}' \tilde{\mathbf{D}}'^{-\frac{1}{2}}) \mathbf{H}'^{(K- 1)}$ we have the compact expression:
$
    \mathbf{H}'^{(K)} \triangleq \sigma(\tilde{\mathbf{H}}'^{(K)} \mathbf{\Theta}^{(K)})
    \label{eq:coarse_gcn_layer}
$.
Note that the dimensions of the parameter matrix, $\mathbf{\Theta}^{(K)}$, of a coarse graph convolution layer is not dependent on the partition $\mathcal{P}$, but rather on the dimensions of the original node attribute matrix $\mathbf{X}$ and can thus be applied to \eqref{eq:gcn_layer}. 

%% file: sections/methodology.tex
\section{\shortname{}: \longname{}}

A coarsening algorithm designed for scalable GNN training should: (1) produce a small coarsened graph, and (2) a GNN fit on the coarsened graph should have a similar prediction performance to a GNN fit on the original graph. 
We hypothesize, and empirically verify in \secref{sec:experiments}, that preserving the output of graph convolutions by minimizing the difference in the intermediate node representations computed for a GCN layer 
and a coarse graph convolution layer 
produces good coarsenings for scalable training of spectral-based GNNs.
In this section, we formalize the notion of preserving the output of graph convolutions with a combinatorial optimization problem.
We then introduce two coarsening methods: \longname{} (\shortname{}) and a highly scalable variant, A-\shortname{}, both approximately solving the proposed optimization problem. 

\subsection{Convolution Matching Objective}
Preserving the output of GCN graph convolution for a given graph signal $\mathbf{x}$ and parameterized filter $\mathbf{g}_{\theta}$ is formalized by the following problem:
{
\begin{align}
    \argmin_{\mathbf{P} \in \mathbb{P}(n, n')} &
    \Vert \theta \mathbf{P} (\tilde{\mathbf{D}}'^{-\frac{1}{2}} \tilde{\mathbf{A}}' \tilde{\mathbf{D}}'^{-\frac{1}{2}}) \mathbf{x}' - \theta (\tilde{\mathbf{D}}^{-\frac{1}{2}} \tilde{\mathbf{A}} \tilde{\mathbf{D}}^{-\frac{1}{2}}) \mathbf{x} \Vert_{1}^{1}.  
\label{eq:coarsening_objective_1}
\end{align}
}%
In words, we aim to find a partition matrix that minimizes the sum of the $L_1$ distances between the node representations obtained via the output of a single graph convolution on the original and coarsened graph.
The parameter $\theta$ acts as a positive scalar multiple in our objective and therefore minimizing the difference in unscaled GCN convolution operation is equivalent.

We generalize the objective in \eqref{eq:coarsening_objective_1} to multi-dimensional graph signals by formulating a multi-objective problem.
Specifically, we equally weigh the difference in the GCN convolution operation for each component of the graph signal to define a linear scalarization of the multi-objective problem: 
{
\begin{align}
    \argmin_{\mathbf{P} \in \mathbb{P}(n, n')} &
    \sum_{i = 1}^{f} \Vert \mathbf{P} (\tilde{\mathbf{D}}'^{-\frac{1}{2}} \tilde{\mathbf{A}}' \tilde{\mathbf{D}}'^{-\frac{1}{2}}) \mathbf{x}_{i}' - (\tilde{\mathbf{D}}^{-\frac{1}{2}} \tilde{\mathbf{A}} \tilde{\mathbf{D}}^{-\frac{1}{2}}) \mathbf{x}_{i} \Vert_{1}^{1}, 
    \label{eq:coarsening_objective_2}
\end{align}
}%
where $\mathbf{x}_{i}$ and $\mathbf{x}'_{i}$ are the $i'th$ components of the $f$-dimensional graph signals $\mathbf{X}$ and $\mathbf{X}'$.
The resulting objective ensures that when fitting GNN parameters using the coarsened graph, the parameters are trained to operate on a matrix that is close to the original.
It is important to note that the coarsening problem formulated in \eqref{eq:coarsening_objective_2} does not restrict the partitioning to preserve connections in the original graph, i.e., two nodes that are distant or even existing in disconnected components of the original graph may be merged into a single supernode.

\begin{figure*}[t!]
    \centering
    \includegraphics[width=0.9\textwidth]{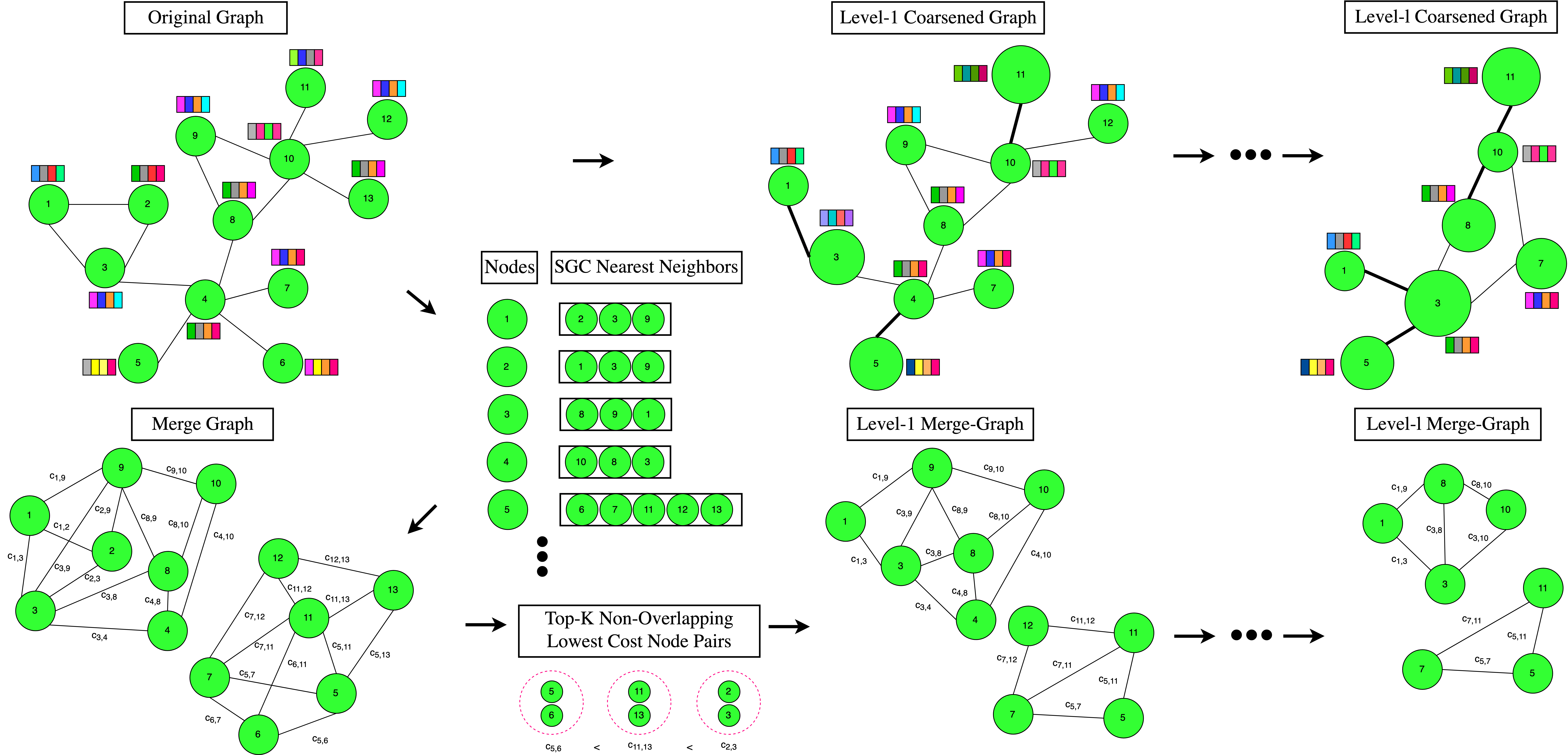}
    \caption{
    Illustration of the \shortname{} algorithm. 
    Nodes with similar SGC embeddings obtained from the original graph are first connected in a \textit{merge-graph}.
    The cost $c_{i, j}$ of merging every pair of nodes $(i, j)$ in the merge-graph is then computed and a set of lowest-cost node pairs are merged into supernodes.
    This process is repeated until the desired coarsening ratio is reached.
    }    \label{fig:scalingGNNsViaGraphCoarsening}
    \vspace{-0.4cm}
\end{figure*}

\subsection{\shortname{}}

\begin{algorithm}[t]
    \footnotesize
    \SetKwComment{Comment}{//}{}
    \SetKwInOut{Input}{input}
    \SetKwInOut{Output}{output}
    
    \caption{\shortname{}} \label{alg:conv_matching}
    
    \Input{Graph $G = (\mathcal{V}, \mathcal{E}, \mathbf{X})$, Ratio $r$, Merges per level $\texttt{k}$}
    \Output{Coarsened graph $G' = (\mathcal{V}', \mathcal{E}', \mathbf{X}')$}
   
    
    $G' = (\mathcal{V}', \mathcal{E}', \mathbf{X}') \gets G = (\mathcal{V}, \mathcal{E}, \mathbf{X});
    \quad \mathbf{P} \gets \mathbf{I}; \quad \tilde{\mathbf{H}}' \gets (\tilde{\mathbf{D}}^{-\frac{1}{2}} \tilde{\mathbf{A}} \tilde{\mathbf{D}}^{-\frac{1}{2}}) \mathbf{X};$
    
    
    $\texttt{candidates} \gets \texttt{CandidateSupernodes}(G');$\\
    
    $\texttt{supernode\_costs} \gets \texttt{ComputeCosts}(G', \tilde{\mathbf{H}}', \texttt{candidates});$
    
    
    \While{$\vert \mathcal{V}' \vert > r \cdot \vert \mathcal{V} \vert$}{

        
        $G', \mathbf{P}, \tilde{\mathbf{H}}', \texttt{supernodes} \gets \texttt{Merge}(G', \texttt{TopKNonOverlap}(\texttt{supernode\_costs}, \texttt{k}));$
        
        
        
        $\texttt{supernode\_costs} \gets \texttt{ComputeCosts}(G', \tilde{\mathbf{H}}', \texttt{Neighborhood}(\texttt{supernodes}));$
    }
\end{algorithm}

A brute-force approach solving \eqref{eq:coarsening_objective_2} by computing the cost of all partitionings of the nodes is an intractable procedure as the number of partitionings grows combinatorially with the number of nodes.
We therefore take a bottom-up hierarchical agglomerative clustering approach with \shortname{} to find an approximate solution.
\shortname{} is outlined in Algorithm~\ref{alg:conv_matching} (and Algs.~\ref{alg:candidate_supernodes}-\ref{alg:supernode_cost} in the Appendix) and illustrated in Figure~\ref{fig:scalingGNNsViaGraphCoarsening}.
In short, \shortname{}  proceeds by first, in \step $1$, computing the intermediate node representation obtained via a coarse graph convolution (\eqref{eq:coarse_gcn_layer}) and creating an initial set of candidate node pairs, or supernodes. 
Then, \step $2$ computes a cost for each pair measuring the change in the objective value of \eqref{eq:coarsening_objective_2} caused by creating the supernode, i.e., the change in the GCN convolution output.
Finally, \step $3$ of \shortname{} finds a number of lowest-cost node pairs, merges them, and finds new node pair candidates and costs.
This process is repeated until the desired coarsening ratio is reached.
As a hierarchical approach, ConvMatch produces multiple levels of coarsening, i.e., we refer to a the graph after $\ell$ passes of the ConvMatch algorithms a level-$\ell$ coarsened graph.
In the following subsections, we describe the processes for generating candidate supernodes, computing supernode costs, and finally merging nodes.


\subsubsection{\step $1$: Candidate Supernodes}
Considering all $\sim n^2$ node pairs as candidate supernodes is infeasible for large scale graphs with millions of nodes and edges.
We therefore only look at a subset of all possible pairs that captures both attribute and structural similarities between nodes.
Specifically, to generate the initial set of candidate supernodes \shortname{} pairs nearest neighbors in the embedding space of a trivially parameterized ($\mathbf{\Theta} = \mathbf{I}$) $K$-layer SGC network:
$
    \mathbf{H}_{SGC}^{(K)} = (\tilde{\mathbf{D}}^{-\frac{1}{2}} \tilde{\mathbf{A}} \tilde{\mathbf{D}}^{-\frac{1}{2}})^{K} \mathbf{X} \mathbf{I}
    \label{eq:trivial_sgc_network}
$.
This embedding is the output of $K$ recursive applications of a GCN convolution \eqref{eq:gcn_convolution}, the very operation we are aiming to preserve. 
The supernode candidate set defines the \emph{merge-graph}: $G_{merge} = (\mathcal{V}', \mathcal{E}_{merge})$, where, initially, $\mathcal{V}' = \mathcal{V}$ and $\mathcal{E}_{merge}$ is the set of edges connecting the generated node pairs.
The embedding step has a computational time complexity of $O(K \cdot d_{avg} \cdot \vert \mathcal{V} \vert)$, where $K$ is the depth of the SGC network being used and $d_{avg}$ is the average degree of nodes in the graph.
Note that computing the embedding is embarrassingly parallelizable.
This step is illustrated in \figref{fig:scalingGNNsViaGraphCoarsening} as the initial merge graph creation.
See \appref{app:candidate_supernodes} for a more detailed description and algorithm.

\subsubsection{\step $2$: Supernode Cost Computation}
Each edge connecting two supernodes, $ u, v \in \mathcal{V}' $, in the merge-graph, $G_{merge} = (\mathcal{V}', \mathcal{E}_{merge})$, is associated with a cost quantifying the objective value in \eqref{eq:coarsening_objective_2} for a partitioning that merges the incident supernodes.
Let $\mathbf{P}_{(u, v)}$ be the partition matrix merging supernodes $u$ and $v$.
Moreover, let $ \tilde{\mathbf{H}}^{(1)}_{(l)} $ and $ \tilde{\mathbf{H}}^{(1)}_{(l, \mathbf{P}_{(u, v)})} $ represent the coarse graph convolution node representations obtained before and after applying the partitioning $ \mathbf{\mathbf{P}} $ at level $l$, respectively.
Then, the cost of merging two supernodes is:
{
\begin{align}
    cost(u, v) \triangleq \Vert \mathbf{P}_{(u, v)} \tilde{\mathbf{H}}^{(1)}_{(l, \mathbf{P}_{(u, v)})} - \tilde{\mathbf{H}}^{(1)}_{(l)} \Vert_{1}^{1}.
    \label{eq:merge_cost}
\end{align}
}%
A scalable algorithm and an illustration of an instance of the supernode cost computation is provided in \appref{app:supernode_costs}.
Computing the cost of merging two nodes, $u$ and $v$, exactly as it is defined in \eqref{eq:merge_cost} has a time complexity of $O(d_{u} + d_{v})$, where $d_{u}$ and $d_{v}$ are the degrees of $u$ and $v$, respectively.
This is a result of the fact that merging nodes $u$ and $v$ effects the representation of each neighbor of $u$ and $v$.
In \figref{fig:scalingGNNsViaGraphCoarsening}, edges in the merge graph are attributed with this cost.
Caching techniques for scaling the evaluation of \eqref{eq:merge_cost} are described in \appref{app:caching_node_summation_terms}.

\textbf{A-\shortname{}.} 
Motivated by the following result, we propose A-\shortname{}, an approximation of the supernode cost computation that yields significant improvements in graph summarization time.
\begin{theorem}
The following is a tight upper bound on \eqref{eq:merge_cost}
{\footnotesize
\begin{align}
    cost&(u, v) \leq \Vert \tilde{\mathbf{H}}^{(1)}_{(l)}[u] - \mathbf{P}_{(u, v)} \tilde{\mathbf{H}}^{(1)}_{(l, \mathbf{P}_{(u, v)})}[(u, v)] \Vert_{1}^{1} + \Vert \tilde{\mathbf{H}}^{(1)}_{(l)}[v] - \mathbf{P}_{(u, v)} \tilde{\mathbf{H}}^{(1)}_{(l, \mathbf{P}_{(u, v)})}[(u, v)] \Vert_{1}^{1} \nonumber\\ 
    &+ \Vert \tilde{\mathbf{x}}_{(u, v)} - \tilde{\mathbf{x}}_{u} \Vert_1^1 \sum_{i \in \mathcal{N}(\{u\})} \frac{a_{u i}}{\sqrt{(d_{i} + \vert C_{i} \vert)}} + \Vert \tilde{\mathbf{x}}_{(u, v)} - \tilde{\mathbf{x}}_{v} \Vert_1^1 \sum_{i \in \mathcal{N}(\{v\})} \frac{a_{v i}}{\sqrt{(d_{i} + \vert C_{i} \vert)}},
\end{align}
}%
where $\tilde{\mathbf{x}}_{i} \triangleq \frac{\mathbf{x}_{i}}{\sqrt{(d_{i} + \vert C_{i} \vert)}}$ are the normalized features for supernode $i$. 
The bound is satisfied with equality when $ \mathcal{N}(u) \cap \mathcal{N}(v) = \emptyset $.
\label{thm:merge_cost_upper_bound}
\end{theorem}

The proof of \theoremref{thm:merge_cost_upper_bound} is provided in \appref{app:cost_approximation}.
We use this bound as an approximation of the cost of merging two nodes in A-\shortname{}. 
This approximation allows the cost of merging two nodes to be a function of properties local to the two nodes being considered, making the cost computation fast and highly scalable.
More formally, the time complexity of computing the approximate cost of merging two nodes, $u$ and $v$, is a constant, $O(1)$, operation.

\subsubsection{\step $3$: Node Merging}

At every level of coarsening, \shortname{} simultaneously merges the top-$k$ non-overlapping lowest-cost candidate supernodes.
For both the coarsened graph and merge-graph, when supernodes $u$ and $v$ are merged to create a new supernode, the new supernode is connected to every neighbor of $u$ and $v$.
Furthermore, the edges connecting supernodes in the resulting coarsened graph are weighted by the number of edges connecting nodes in the two incident supernodes.
Moreover, the features of the supernodes are a weighted average of the features of the nodes being merged and in node classification settings, the node label used for training is the majority label of nodes in a supernode.
In addition, the cost of a subset of node pairs connected by an edge in the merge-graph must be updated after a merge.
More formally, the time complexity of merging two nodes $u$ and $v$ is roughly $O(d^{merge}_{avg} \cdot (d_{u} + d_{v}))$, where $d^{merge}_{avg}$ is the average degree of nodes in the merge-graph.
This process is also highly parallelizable.
In \figref{fig:scalingGNNsViaGraphCoarsening}, node merging is performed to obtain higher levels of coarsening, i.e., level-$l$ merge- and coarsened-graphs.
Details on a highly scalable merging procedure are provided in \appref{app:merging_nodes} and a scalable cost computation and update in \appref{app:cost_approximation}.

%% file: sections/evaluation.tex
\section{Experiments}
\label{sec:experiments}

We perform experiments to answer the following research questions: 
(\textbf{RQ1}) At varying coarsening ratios, how do our \shortname{} variants compare to the baselines in terms of summarization time, as well as GNN training runtime and memory requirements?
(\textbf{RQ2}) How effective are the GCNs trained on graphs summarized by our \shortname{} variants (vs.\ baselines) in downstream node classification and link prediction tasks?
(\textbf{RQ3}) What is the effect of the number of merges $k$ at each level of coarsening in \shortname{} on the summarization time and downstream task performance?
All reported results are fully reproducible, with code and data available at: \url{github.com/amazon-science/convolution-matching}.

\begin{wraptable}{r}{0.55\textwidth}
\vspace{-0.5cm}
    \centering
    \caption{
    Table of dataset statistics and task (NC: node classification; LP: link prediction).
    }
    \scalebox{0.7}{
    \begin{tabular}{lcccc}
        \toprule
        \textbf{Dataset} & \textbf{Task} & \textbf{Nodes} & \textbf{Edges} & \textbf{Features} \\
        \midrule
        Citeseer & NC / LP & $3,327$ & $4,732$ & $3,703$ \\
        Cora & NC / LP & $2,708$ & $5,429$ & $1,433$ \\
        OGBNArxiv & NC & $169,343$ & $1,166,243$ & $128$ \\
        OGBLCollab & LP & $235,868$ & $1,285,465$ & $128$ \\
        OGBLCitation2 & LP & $2,927,963$ & $30,561,187$ & $128$ \\
        OGBNProducts & NC & $2,449,029$ & $61,859,140$ & $100$ \\
        \bottomrule
    \end{tabular}
    }
    \label{tab:datasets}
    \vspace{-0.2cm}
\end{wraptable}
\noindent \textbf{Datasets.} In our experiments, we use six datasets summarized in \tabref{tab:datasets}.
Citeseer and Cora are citation networks which we use for both node classification (NC) and link prediction (LP) tasks \citep{sen:ai08}.
Additionally, we use four datasets from the Open Graph Benchmark (OGB) \citep{hu:nips20}.
OGBNArxiv (Arxiv) and OGBLCitation2 (Cit2) are also citation networks and are curated for testing NC and LP performance, respectively. 
OGBLCollab (Coll) is a collaboration network with the LP task of ranking true collaborations higher than false collaborations.
Finally, OGBNProducts (Prod) is a product co-purchasing network with the NC task of predicting product categories.
Prediction performance for NC tasks is measured using accuracy.
The prediction performance for LP tasks is measured using AUC on Citeseer and Cora, Hits@50 on Coll, and MRR on Cit2.

\par {\textbf{Baselines}.}
We additionally evaluate the performance of six baselines:
(a)~three coreset methods: \emph{Random Node Sampling} (RS), \emph{Herding} \citep{welling:icml09}, and \emph{KCenter} \citep{sener:iclr18};
(b)~two graph condensation methods: \emph{graph condensdation} (GCond) \citep{jin:iclr22} and \emph{one-step gradient matching} (DosCond) \citep{jin:kdd22};
and (c)~one coarsening method: \emph{Variation Neighborhoods} (VN) \citep{loukas:jmlr19, huang:kdd21}.
For implementation and hyperparameter details see \appref{app:extended_evaluation}.

\par{\textbf{GCN Architectures and Hyperparameters}}.
All experiments are performed using a GCN model.
We give the details of the GCN architectures and hyperparameters for summarization baselines and \shortname{} and A-\shortname{} in \appref{app:extended_evaluation}. 
The merge batch sizes of our algorithm is fixed for each dataset for experiments in \secref{sec:exp_runtime_memory} and \secref{sec:exp_prediction_performance} and is also provided in \appref{app:extended_evaluation}.

\subsection{(RQ1) Runtime and Memory Efficiency}
\label{sec:exp_runtime_memory}

\input{figures/graphSummarizationTimes}
 
First, we evaluate the efficiency of our proposed \shortname{} variants and the baseline graph summarization algorithms, as well as the training time and memory efficiency of the GNNs trained on the resultant graph summaries.


\par{\textbf{Graph Summarization Time.}} We compare the graph summarization time of \shortname{} and A-\shortname{} to baselines at varying coarsening ratios for each dataset and task.
The complete set of results are provided in \appref{app:extended_evaluation}.
The average time across 5 rounds of summarization for Cora, OGBNArxiv, and OGBNProducts are shown in \figref{fig:graphSummarizationTimes}.

First, we observe for Cora A-\shortname{} is over $5 \times$ faster than \shortname{}.
For this reason we chose to only run A-\shortname{} on the larger OGB datasets.
A-\shortname{} is consistently faster than all other baseline graph summarization methods on the larger OGB datasets.
The VN baseline is faster than A-\shortname{} in Cora, however we empirically demonstrate this method struggles to scale to larger graphs 
(e.g., it timed out after 24 hours on the OGBNProducts dataset).
A-\shortname{} is faster than the condensation and coreset baselines as it does not compute gradients of a GNN model with the full graph during summarization.
Finally, we note that the coarsening methods have an additional advantage of being bottom-up multi-level approaches and thus the time required to reach the coarsening ratio $r=0.1 \%$ includes the time required to reach the ratio $r=1 \%$ and $r=10 \%$ and so on.
On the other hand, creating the synthetic graphs with the condensation methods for two different ratios is two separate procedures and the work to reach one ratio is not obviously usable to reach another. 
In practice this property could be leveraged in GNN learning curriculums or hyperparameter exploration.

\par{\textbf{GNN Training Runtime and Memory Efficiency.}} 
We examine the amount of GPU memory used and the time required to complete a fixed number of training epochs for each dataset at varying coarsening ratios.
\tabref{tab:training_times} shows the average total time and maximum amount of GPU memory used across 5 rounds of training on a graph coarsened using A-\shortname{}.
\tabref{tab:training_times} shows there is a significant decrease in the amount of GPU memory and time required to complete training on a coarsened graph. 
The results are most notable on the largest datasets: OGBLCitation2 and OGBNProducts, where the amount of memory required to compute the batch gradient for the GCN exceeded the $120 GB$ of GPU memory available on our machine.

\input{sections/tables/training_times}

\subsection{(RQ2) Downstream Task Prediction Performance}
\label{sec:exp_prediction_performance}

\input{sections/tables/performance}


We now compare the prediction performance of GCNs trained on graphs summarized using \shortname{}, A-\shortname{}, and baseline summarization methods at varying coarsening ratios.
Link prediction and node classification performance of the trained GCNs are reported in \tabref{tab:link_prediction_performance} and \tabref{tab:node_classification_performance}, respectively.
\shortname{} or A-\shortname{} is consistently among the top three performing summarization methods for both tasks.
Furthermore, A-\shortname{} achieves the best overall performance, as indicated by the lowest average rank.

\tabref{tab:link_prediction_performance} shows \shortname{} and A-\shortname{} are significantly better at creating summarized graphs for training a GCN for link prediction compared to alternative summarization methods. 
Notably, GCN's trained on \shortname{} and A-\shortname{} summarized graphs achieve up to $90\%$ of the link prediction performance at a coarsening ratio of $r=1\%$ in Citeseer and Cora, respectively. 
Moreover, A-\shortname{} achieves a nearly 2 $\times$ improvement over the best performing baseline in Cit2 at $r=0.1\%$ and over a 20\% point improvement at $r=1\%$.
A possible explanation for this is that \shortname{} and A-\shortname{} merge nodes that are equivalent or similar with respect to the GCN convolution operation, which captures both nodes attributes and structural properties.
For this reason the summarized graph contains supernodes that can be used to create good positive and negative link training examples.

\tabref{tab:node_classification_performance} shows Herding and KCenter methods perform well for the node classification task on the larger OGB datasets. 
These methods do however have a trade off as they initially fit a GNN using the complete graph to obtain node embeddings resulting in a slower summarization time, as shown in the previous section.
Condensation methods perform extremely well on Citeseer and Cora, however they have difficulty creating larger summarized graphs hitting out of memory errors on the OGB datasets because they compute a full gradient using the original graph.
Furthermore, the results we find for GCond on OGBNArxiv differ from those reported in \citenoun{jin:kdd22} \cite{jin:kdd22} as the GCN architecture in the summarizer and the GCN being trained are not exactly matched. 
The authors mention this behavior in their appendix section C.5.
Finally, we observe A-\shortname{} is the most reliable summarizer for node classification with the best average rank of $2.3$.
The next best method in terms of average rank is Herding at $3.3$.

\subsection{\textbf{(RQ3)} Ablation for \shortname{} Merge Batch Size}

\input{sections/tables/merge_batch_ablation}

First, we analyze the effect the number of node pairs simultaneously merged at each level of coarsening in \shortname{} has on the summarization time and prediction performance of the proposed approach.
We summarize the training graph for each of the 6 datasets to the coarsening ratio $r=1.0\%$ and train a GCN using the summarized graph.
The summarization time in seconds and prediction performance are reported in \tabref{tab:merge_batch_ablation} for both link prediction and node classification.
We find increasing the merge batch size has a limited effect on the prediction performance across all datasets and for both NC and LP tasks.
However, the summarization time improves considerably.

%% file: figures/graphSummarizationTimes.tex
\begin{figure*}
\begin{subfigure}{.33\textwidth}
  \centering
  \includegraphics[width=.9\linewidth]{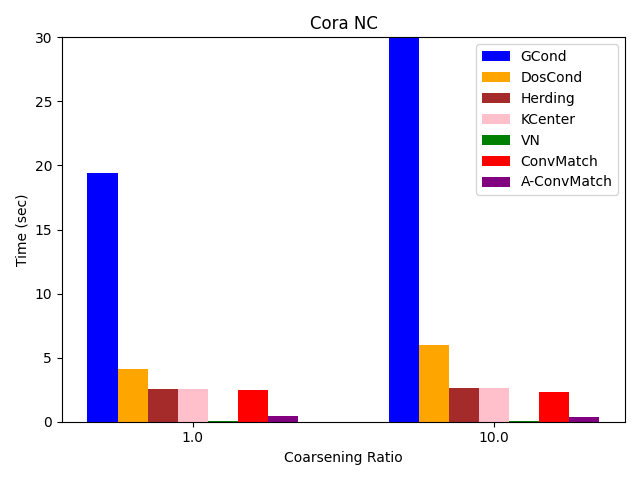}
  \label{fig:coraNCGraphSummarizationTime}
\end{subfigure}%
\begin{subfigure}{.33\textwidth}
  \centering
  \includegraphics[width=.9\linewidth]{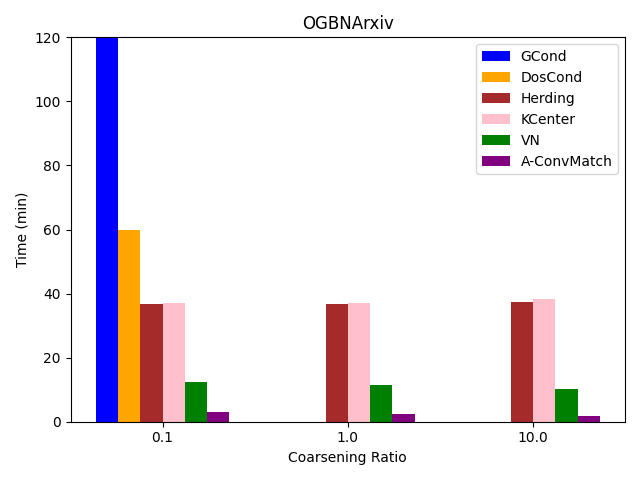}
  \label{fig:OGBNArxivGraphSummarizationTime}
\end{subfigure}%
\begin{subfigure}{.33\textwidth}
  \centering
  \includegraphics[width=.9\linewidth]{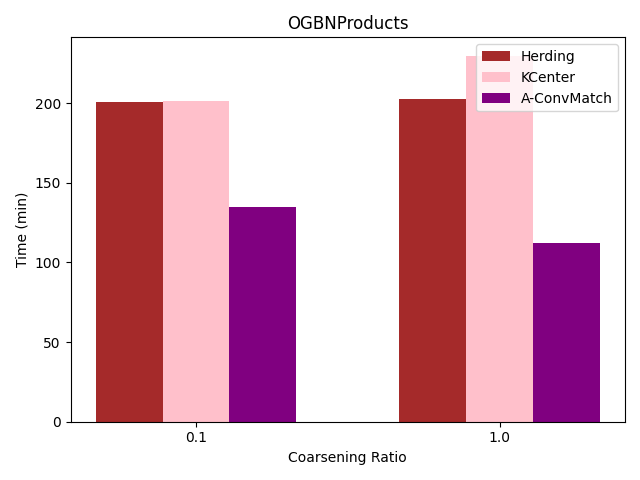}
  \label{fig:OGBNProductsGraphSummarizationTime}
\end{subfigure}
\caption{Plots of graph summarization times at multiple coarsening ratios for Cora, Arxiv, and Prod datasets.
\shortname{} and A-\shortname{} are fast summarization algorithms. 
}
\label{fig:graphSummarizationTimes}
\vspace{-0.4cm}
\end{figure*}

%% file: sections/tables/training_times.tex
\begin{table}[tb!]
    \centering    
    \caption{
    Average time, rounded to the nearest minute, and GPU memory, rounded to the nearest GB, required to complete all training epochs at varying coarsening ratios.
    }
    \label{tab:training_times}
    \begin{subtable}[t]{0.45\textwidth}
        \centering
        \caption{Link Prediction}
        \scalebox{0.7}{
        \begin{tabular}{lr@{\hskip 0.5cm}r@{\hskip 0.4cm}r}
        \toprule
            \textbf{Dataset} & \textbf{Ratio} & \textbf{Time (min)} & \textbf{Memory (GB)} \\
        \midrule
            \multirow{3}{*}{Citeseer} 
            & $1$ & $ 3 $ & $ 2 $ \\
            & $10$ & $ 3 $ & $ 3 $ \\
            & $100$ & $ 10 $ & $ 12 $ \\
        \midrule
            \multirow{3}{*}{Cora} 
            & $1$ & $ 4 $ & $ 1 $ \\
            & $10$ & $ 4 $ & $ 1 $ \\
            & $100$ & $ 12 $ & $ 7 $ \\
        \midrule
            \multirow{4}{*}{Coll} 
            & $0.1$ & $ 6 $ & $ 9 $ \\
            & $1$ & $ 15 $ & $ 9 $ \\
            & $10$ & $ 40 $ & $ 12 $ \\
            & $100$ & $ 210 $ & $ 70 $ \\
        \midrule
            \multirow{3}{*}{Cit2} 
            & $0.1$ & $ 60 $ & $ 22 $ \\
            & $1$ & $ 660 $ & $ 40 $ \\
            & $100$ & $ >1440 $ & $ >120 $ \\
        \bottomrule
        \end{tabular}
        }
    \end{subtable}
    \begin{subtable}[t]{0.45\textwidth}
        \centering
        \caption{Node Classification}
        \scalebox{0.7}{
        \begin{tabular}{lr@{\hskip 0.5cm}r@{\hskip 0.4cm}r}
        \toprule
            \textbf{Dataset} & \textbf{Ratio} & \textbf{Time (min)} & \textbf{Memory (GB)} \\
        \midrule
            \multirow{3}{*}{Citeseer} 
            & $1$ & $ 4 $ & $ 1 $ \\
            & $10$ & $ 5 $ & $ 1 $ \\
            & $100$ & $ 5 $ & $ 1 $ \\
        \midrule
            \multirow{3}{*}{Cora} 
            & $1$ & $ 2 $ & $ 1 $ \\
            & $10$ & $ 4 $ & $ 1 $ \\
            & $100$ & $ 4 $ & $ 1 $ \\
        \midrule
            \multirow{4}{*}{Arxiv} 
            & $0.1$ & $ 5 $ & $ 7 $ \\
            & $1$ & $ 10 $ & $ 7 $ \\
            & $10$ & $ 20 $ & $ 8 $ \\
            & $100$ & $ 170 $ & $ 20 $ \\
        \midrule
            \multirow{3}{*}{Prod} 
            & $0.1$ & $ 60 $ & $ 38 $ \\
            & $1$ & $ 150 $ & $ 45 $ \\
            & $100$ & $ >1440 $ & $ >120 $ \\
        \bottomrule
        \end{tabular}
        }
    \end{subtable}
    \vspace{-0.3cm}
\end{table}

%% file: sections/tables/performance.tex
\begin{table}[tb]
    \centering
    \caption{Link prediction and node classification performance at varying coarsening levels. 
    The top three performing scores are highlighted as: \colorbox{orange!50}{First}, \colorbox{orange!25}{Second}, \colorbox{orange!12}{Third}. 
    Average ranks are reported for methods that were ran on all datasets and coarsening ratios.
    Our A-\shortname{} approach performs the best across datasets and coarsening ratios, as indicated by the lowest average rank in both link prediction and node classification tasks (1.4 and 2.5, resp.).
    } 
\begin{subtable}[t]{\textwidth}
    \centering
    \caption{Link Prediction}
    \scalebox{0.6}{
    \begin{tabular}{cr@{\hskip 0.5cm}cccc@{\hskip 0.5cm}cc}
    \toprule
         & & \multicolumn{6}{c}{\textbf{Sumarizer}} \\
        \cmidrule{3-8}
         & \textbf{Ratio} & \textbf{RS} & \textbf{GCond} & \textbf{DosCond} & \textbf{VN} & \textbf{\shortname{}} (ours)  & \textbf{A-\shortname{}} (ours)  \\
    \midrule
        \multirow{3}{*}{\rotatebox{90}{Citeseer}}
        & $1\%$ & $ 67.41 \pm 1.26 $ & $ 65.01 \pm 1.09 $ & $ 63.54 \pm 6.36 $ & \cellcolor{orange!12} $ 80.26 \pm 4.51 $ & \cellcolor{orange!50} $ \mathbf{87.72 \pm 1.38} $ & \cellcolor{orange!25} $ 87.68 \pm 1.24 $ \\
        & $10\%$ & $ 69.11 \pm 2.09 $ & $ 72.65 \pm 6.78$ & $ 53.06 \pm 12.28 $ & \cellcolor{orange!25} $ 88.55 \pm 0.77 $ & \cellcolor{orange!50} $ \mathbf{90.43 \pm 0.49} $ & \cellcolor{orange!12} $ 88.41 \pm 0.97 $ \\
    \cline{3-8}
        & $100\%$ & \multicolumn{6}{c}{\cellcolor{Gray} $ \mathbf{91.57 \pm 0.55} $} \\
    \midrule
    \multirow{3}{*}{\rotatebox{90}{Cora}}
        & $1\%$ & $ 70.91 \pm 3.67 $ & $ 67.08 \pm 0.64 $ & \cellcolor{orange!12} $ 71.60 \pm 0.35 $ & \cellcolor{orange!25} $ 75.19 \pm 3.01 $ & $ 66.33 \pm 2.35 $ & \cellcolor{orange!50} $ \mathbf{78.63 \pm 1.55} $ \\
        & $10\%$ & $ 72.84 \pm 2.15 $ & $ 66.19 \pm 0.39 $ & $ 60.21 \pm 0.52 $ & \cellcolor{orange!12} $ 81.81 \pm 2.06 $ & \cellcolor{orange!50} $ \mathbf{83.51 \pm 1.01} $ & \cellcolor{orange!25} $ 83.30 \pm 1.45 $ \\
    \cline{3-8}
        & $100\%$ & \multicolumn{6}{c}{\cellcolor{Gray} $ \mathbf{85.04 \pm 0.68} $} \\
    \midrule
    \multirow{4}{*}{\rotatebox{90}{Coll}}
        & $.1\%$ & \cellcolor{orange!12} $ 7.22 \pm 0.97 $ & \cellcolor{orange!25} $ 7.61 \pm 0.41  $ & $ 1.15 \pm 0.88 $ & $ 6.60 \pm 1.47 $ & - & \cellcolor{orange!50} $ \mathbf{9.14 \pm 2.07} $ \\
        & $1\%$ & \cellcolor{orange!25} $ 11.41 \pm 0.35 $ & OOM & OOM & \cellcolor{orange!12} $ 5.45 \pm 1.95 $ & - & \cellcolor{orange!50} $ \mathbf{26.25 \pm 1.84} $ \\
        & $10\%$ & \cellcolor{orange!12} $ 15.80 \pm 3.11 $ & OOM & OOM & \cellcolor{orange!25} $ 24.33 \pm 1.61 $ & - & \cellcolor{orange!50} $ \mathbf{35.94 \pm 1.86} $ \\
    \cline{3-8}
        & $100\%$ & \multicolumn{6}{c}{\cellcolor{Gray} $ \mathbf{44.08 \pm 0.94} $} \\
    \midrule
    \multirow{3}{*}{\rotatebox{90}{Cit2}}
        & $.1\%$ & \cellcolor{orange!12} $ 9.86 \pm 0.01 $ & OOM & OOM & \cellcolor{orange!25} $ 10.43 \pm 2.71 $ & - & \cellcolor{orange!50} $ \mathbf{30.42 \pm 0.63} $ \\
        & $1\%$ & \cellcolor{orange!12} $ 9.87 \pm 0.01 $ & OOM & OOM & \cellcolor{orange!25} $ 38.95 \pm 4.83 $ & - & \cellcolor{orange!50} $ \mathbf{60.34 \pm 1.15} $ \\
    \cline{3-8}
         & $100\%$ &  \multicolumn{6}{c}{\cellcolor{Gray} $ \mathbf{84.74 \pm 0.00} $} \\
    \midrule
        \multicolumn{2}{c}{\textbf{Avg. Rank}} & $3.4$ & $-$ & $-$ & $2.6$ & $-$ & $\textbf{1.4}$ \\
    \bottomrule
    \end{tabular}
    }
    \label{tab:link_prediction_performance}
\end{subtable}
\begin{subtable}[t]{\textwidth}
    \centering
    \caption{Node Classification}
    \scalebox{0.6}{
    \begin{tabular}{cr@{\hskip 0.5cm}cccccc@{\hskip 0.5cm}cc}
    \toprule
         & & \multicolumn{8}{c}{\textbf{Sumarizer}} \\
        \cmidrule{3-10}
         & \multirow{2}{*}{\textbf{Ratio}} & \multirow{2}{*}{\textbf{RS}} & \multirow{2}{*}{\textbf{KCenter}} &
         \multirow{2}{*}{\textbf{Herding}} &
         \multirow{2}{*}{\textbf{GCond}} & \multirow{2}{*}{\textbf{DosCond}} & \multirow{2}{*}{\textbf{VN}} & \textbf{\shortname{}}  & \textbf{A-\shortname{}}  \\
         & & \multicolumn{6}{c}{} & (ours) & (ours)\\
    \midrule
        \multirow{3}{*}{\rotatebox{90}{Citeseer}}
        & $1\%$ & $ 19.22 \pm 6.85 $ & $ 59.02 \pm 2.25 $ & $ 62.80 \pm 1.33 $ & \cellcolor{orange!25} $ 68.22 \pm 2.08 $ & \cellcolor{orange!50} $ \mathbf{71.36 \pm 1.27} $ & $ 35.92 \pm 3.76 $ & $ 65.40 \pm 4.12 $ & \cellcolor{orange!12} $ 67.68 \pm 2.87 $ \\
        & $10\%$ & $ 28.00 \pm 9.15 $ & $ 54.70 \pm 3.62 $ & $ 53.5 \pm 1.68 $ & \cellcolor{orange!50} $ \mathbf{70.10 \pm 3.43} $ & \cellcolor{orange!25} $ 68.76 \pm 0.97 $ & $ 53.88 \pm 7.36 $ & \cellcolor{orange!12} $ 69.84 \pm 1.58 $ & $ 68.46 \pm 1.61 $ \\
        & $100\%$ & \multicolumn{8}{c}{\cellcolor{Gray} $ \mathbf{71.40 \pm 0.35} $} \\
    \midrule
        \multirow{3}{*}{\rotatebox{90}{Cora}}
        & $1\%$ & $ 18.76 \pm 9.67 $ & $ 59.92 \pm 1.67 $ & $ 63.74 \pm 2.24 $ & \cellcolor{orange!25} $ 77.44 \pm 1.90 $ & \cellcolor{orange!50} $ 78.40 \pm 0.83 $ & $ 31.96 \pm 11.02 $ & \cellcolor{orange!12} $ 72.60 \pm 3.11 $ & $ 72.30 \pm 2.90 $ \\
        & $10\%$ & $ 27.52 \pm 9.89 $ & $ 56.12 \pm 7.46 $ & $ 64.10 \pm 2.12 $ & \cellcolor{orange!25} $ 80.02 \pm 0.92 $ & $ 78.64 \pm 1.87 $ & $ 59.22 \pm 4.90 $ & \cellcolor{orange!12} $ 79.82 \pm 0.60 $ & \cellcolor{orange!50} $ 80.12 \pm 1.09 $ \\
        & $100\%$ & \multicolumn{8}{c}{\cellcolor{Gray} $ \mathbf{ 81.02 \pm 0.19 } $} \\
    \midrule
        \multirow{4}{*}{\rotatebox{90}{Arxiv}}
        & $0.1\%$ & $ 42.45 \pm 3.62 $ & \cellcolor{orange!12} $ 49.79 \pm 2.87 $ & \cellcolor{orange!50} $ \mathbf{58.79 \pm 0.90} $ & $ 48.10 \pm 3.27 $ & $ 21.69 \pm 4.30 $ & $ 28.90 \pm 5.79 $ & - & \cellcolor{orange!25} $ 54.82 \pm 2.67 $ \\
        & $1\%$ & $ 59.97 \pm 1.58 $ & \cellcolor{orange!25} $ 61.99 \pm 1.24 $ & \cellcolor{orange!12} $ 61.42 \pm 1.34 $ & OOM & OOM & $ 53.61 \pm 2.76 $ & - & \cellcolor{orange!50} $ \mathbf{64.08 \pm 0.39} $ \\
        & $10\%$ & $ 67.50 \pm 0.63 $ & \cellcolor{orange!50} $ \mathbf{68.25 \pm 0.25} $ & \cellcolor{orange!25} $ 67.75 \pm 0.58 $ & OOM & OOM & $ 67.12 \pm 1.26 $ & - & \cellcolor{orange!12} $ 67.51 \pm 0.47 $ \\
        & $100\%$ & \multicolumn{8}{c}{\cellcolor{Gray} $ \mathbf{ 72.14 \pm 0.18 } $} \\
    \midrule
        \multirow{3}{*}{\rotatebox{90}{Prod}}
        & $0.1\%$ & $ 48.20 \pm 1.57 $ & \cellcolor{orange!12} $ 60.88 \pm 1.31 $ & \cellcolor{orange!50} $ \mathbf{66.14 \pm 0.71} $ & OOM & OOM & TIMEOUT & - & \cellcolor{orange!25} $ 62.36 \pm 0.84 $ \\
        & $1\%$ & $ 66.57 \pm 1.47 $ & \cellcolor{orange!25} $ 69.66 \pm 0.31 $ & \cellcolor{orange!50} $ \mathbf{71.79 \pm 0.45} $ & OOM & OOM & TIMEOUT & - & \cellcolor{orange!12} $ 68.20 \pm 0.36 $ \\
        & $100\%$ & \multicolumn{8}{c}{\cellcolor{Gray} $ \mathbf{ 75.64 \pm 0.00 }^* $} \\
    \midrule
        \multicolumn{2}{c}{\textbf{Avg. Rank}} & $5.9$ & $3.8$ & $3.2$ & - & - & - & - & $\textbf{2.5}$\\
    \bottomrule
    \end{tabular}
    }
    \label{tab:node_classification_performance}
\end{subtable}
\label{tab:prediction_performance}
\vspace{-0.3cm}
\end{table}

%% file: sections/tables/merge_batch_ablation.tex
\begin{table}[tb]
    \centering    
    \caption{
    \shortname{} graph summarization time in seconds and prediction performance at varying merge batch sizes and a coarsening ratio $r=1.0\%$.
    }
    \label{tab:merge_batch_ablation}
    \begin{subtable}[t]{0.45\textwidth}
        \centering
        \caption{Link Prediction}
        \scalebox{0.7}{
        \begin{tabular}{lr@{\hskip 1cm}c@{\hskip 0.6cm}c@{\hskip 0.4cm}c}
        \toprule
            \multirow{2}{*}{\textbf{Dataset}} & \multirow{2}{*}{\textbf{Batch Size}} & \multirow{2}{*}{\textbf{Time (sec)}} & \multicolumn{2}{c}{\textbf{Perf.}} \\
            &&& Valid. & Test\\
        \midrule
            \multirow{3}{*}{Citeseer} 
            & $1$ & $ 178.03 $ & $ 85.42 $ & $ 87.68 $ \\
            & $10$ & $ 37.95 $ & $ 86.78 $ & $ 86.44 $ \\
            & $100$ & $ 8.46 $ & $ 88.19 $ & $ 88.25 $ \\
        \midrule
            \multirow{3}{*}{Cora} 
            & $1$ & $ 105.46 $ & $ 74.21 $ & $ 74.93 $ \\
            & $10$ & $ 25.73 $ & $ 76.20 $ & $ 78.44 $ \\
            & $100$ & $ 4.72 $ & $ 69.98 $ & $ 67.53 $ \\
        \midrule
        \multirow{3}{*}{Coll} 
            & $100$ & $ 777.77 $ & $ 22.59 $ & $ 26.23 $ \\
            & $1,000$ & $ 259.11 $ & $ 22.78 $ & $ 24.63 $ \\
            & $10,000$ & $ 205.53 $ & $ 23.70 $ & $ 26.25 $ \\
        \midrule
            \multirow{2}{*}{Cit2} 
            & $10,000$ & $ 6,555 $ & $ 60.05 $ & $ 60.05 $ \\
            & $100,000$ & $ 4,740 $ & $ 55.21 $ & $ 55.24 $ \\
        \bottomrule
        \end{tabular}
        }
    \end{subtable}
    \begin{subtable}[t]{0.45\textwidth}
        \centering
        \caption{Node Classification}
        \scalebox{0.7}{
        \begin{tabular}{lr@{\hskip 1cm}c@{\hskip 0.6cm}c@{\hskip 0.4cm}c}
        \toprule
            \multirow{2}{*}{\textbf{Dataset}} & \multirow{2}{*}{\textbf{Batch Size}} & \multirow{2}{*}{\textbf{Time (sec)}} & \multicolumn{2}{c}{\textbf{Perf.}} \\
            &&& Valid. & Test.\\
        \midrule
            \multirow{3}{*}{Citeseer} 
            & $1$ & $ 198.84 $ & $ 66.92 $ & $ 66.08 $ \\
            & $10$ & $ 36.78 $ & $ 72.40 $ & $ 68.90 $ \\
            & $100$ & $ 8.03 $ & $ 65.20 $ & $ 62.30 $ \\
        \midrule
            \multirow{3}{*}{Cora} 
            & $1$ & $ 101.74 $ & $ 74.40 $ & $ 74.70 $ \\
            & $10$ & $ 27.16 $ & $ 70.20 $ & $ 72.30 $ \\
            & $100$ & $ 4.89 $ & $ 71.40 $ & $ 71.20 $ \\
        \midrule
        \multirow{3}{*}{Arxiv} 
            & $100$ & $ 624.91 $ & $ 63.94 $ & $ 63.46 $ \\
            & $1,000$ & $ 194.73 $ & $ 62.25 $ & $ 61.07 $ \\
            & $10,000$ & $ 149.91 $ & $ 64.05 $ & $ 64.12 $ \\
        \midrule
            \multirow{3}{*}{Prod} 
            & $10,000$ & $ 11,285 $ & $ 85.95 $ & $ 68.57 $ \\
            & $100,000$ & $ 6,729 $ & $ 86.18 $ & $ 68.20 $ \\
        \bottomrule
        \end{tabular}
        }
    \end{subtable}
    \vspace{-0.3cm}
\end{table}

%% file: sections/conclusion.tex
\section{Conclusion}
\label{sec:conclusion}

We introduced the \shortname{} graph summarization algorithm and a principled approximation, A-\shortname{}, which preserve the output of graph convolution. 
Our methods were empirically proven to produce summarized graphs that can be used to fit GCN model parameters with significantly lower memory consumption, faster training times, and good prediction performance on both node classification and link prediction tasks, a first for summarization for scalable GNN training. 
Notably, our model is consistently a top-performing summarization method and achieves up to 20\% point improvements on link prediction tasks. 
There are exciting next steps to this research, including extending the idea of convolution matching to heterogeneous graphs and developing GNN training algorithms that leverage multiple levels of a graph coarsening.
Moreover, although in this work we focus on motivating our approach and providing a comprehensive evaluation for GCN’s, no change to our coarsening algorithm is necessary to apply it to a different models, for instance a GraphSAGE model may be trained on the condensed graph produced by ConvMatch. 
Future research on applying the same algorithmic framework of ConvMatch but specialized to preserving the operations applied in another GNN model is promising.

%% file: sections/appendix.tex
\section{Appendix}

Here we expand on the theoretical contributions of this work and provide further details on the empirical evaluation presented in \secref{sec:experiments}.
The appendix includes the following sections: Extended Step 1: Candidate Supernodes, Extended Step 2: Computing Supernode Costs, Extended Step 3: Merging Nodes, Caching Node Summation Terms, A Supernode Cost Approximation, and Extended Evaluation.
All reported results are fully reproducible, with code and data available at: \url{github.com/amazon-science/convolution-matching}.

\section{Extended Step 1: Candidate Supernodes}
\label{app:candidate_supernodes}

\begin{algorithm}
    \footnotesize
    \SetKwComment{Comment}{//}{}
    \SetKwInOut{Input}{input}
    \SetKwInOut{Output}{output}
    
    \caption{CandidateSupernodes}\label{alg:candidate_supernodes}
    
    \Input{Graph $G' = (\mathcal{V}', \mathcal{E}', \mathbf{X}')$, Number of neighbors per node $k_1$, Number of neighbors $k_2$}
    \Output{Set of candidate supernodes $\texttt{candidates}$}
    
    $\texttt{candidates} \gets \emptyset$
    
    $\mathbf{H}_{SGC}^{(K)} = (\tilde{\mathbf{D}}^{-\frac{1}{2}} \tilde{\mathbf{A}} \tilde{\mathbf{D}}^{-\frac{1}{2}})^{K} \mathbf{X}$
    
    \textcolor{gray}{\Comment{Compute the exact SGC matches.}}
    $\texttt{candidates} \gets \texttt{candidates} \cup \left \{ (i, j) \,  \vert \, \mathbf{H}_{SGC}^{(K)}[i] = \mathbf{H}_{SGC}^{(K)}[j] \right \}$
    
    \textcolor{gray}{\Comment{Compute the nearest SGC matches for each node.}}
    $\texttt{candidates} \gets \texttt{candidates} \cup \left \{ (i, j) \,  \vert \, j \in \argmin^{k_1}_{j'} \left \Vert \mathbf{H}_{SGC}^{(K)}[i] - \mathbf{H}_{SGC}^{(K)}[j'] \right \Vert_{1}^{1} \right \}$
    
    \textcolor{gray}{\Comment{Compute the nearest SGC matches overall.}}
    $\texttt{candidates} \gets \texttt{candidates} \cup \left \{ (i, j) \,  \vert \, (i, j) \in \argmin^{k_2}_{(i',j')} \left \Vert \mathbf{H}_{SGC}^{(K)}[i'] - \mathbf{H}_{SGC}^{(K)}[j'] \right \Vert_{1}^{1} \right \}$
\end{algorithm}

The initial supernode candidate set generation process is detailed in \algref{alg:candidate_supernodes}.
To ensure every node has the potential to be merged into a supernode, we find the top $k_{nn}$-nearest neighbors for each node and add the node pair to $\mathcal{E}_{merge}$.
Additionally, the nearest $d_{nn}\%$ of all node pairs are added to $\mathcal{E}_{merge}$.
The supernode candidate set defines the \emph{merge-graph}: $G_{merge} = (\mathcal{V}', \mathcal{E}_{merge})$, that is updated throughout the coarsening processes.
Specifically, all nodes from the original graph are initially added to the merge-graph, i.e., initially, $\mathcal{V}' = \mathcal{V}$, then two nodes are connected in $G_{merge}$ if their pair exists in the candidate supernode set.
Details on merging nodes in the coarsened graph and the merge-graph are given in \appref{app:merging_nodes}.

\section{Extended Step 2: Computing Supernode Costs}
\label{app:supernode_costs}

\begin{figure}
    \centering
    \includegraphics[width=0.65\textwidth]{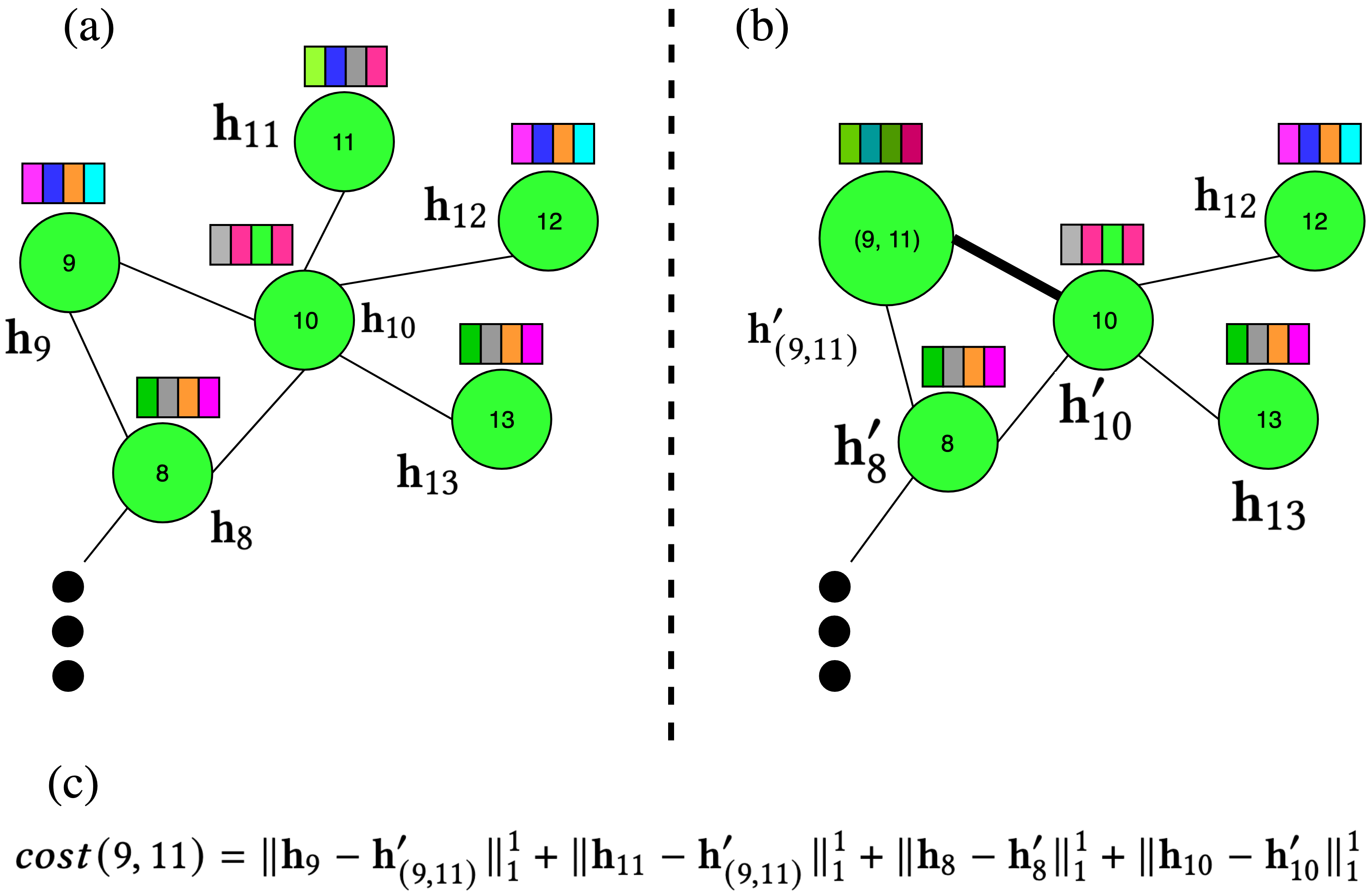}
    \caption{
    (a) A portion of a graph with computed representations $\mathbf{h}_{i}$ for each node $i \in \mathcal{V}$. 
    (b) A portion of a graph with the nodes $9$ and $11$ merged into a supernode. 
    The updated representations of the nodes $8$ and $10$ denoted by $\mathbf{h}'_{8}$ and $\mathbf{h}'_{10}$, respectively.
    The representation of the supernode resulting from the merge is $\mathbf{h}'_{(9, 11)}$.
    (c) The cost of merging the nodes $9$ and $11$ is the sum of the absolute differences in the representations caused by the merge.}
    \label{fig:supernodeCosts}
\end{figure}

\begin{algorithm}
    \small
    \SetKwComment{Comment}{//}{}
    \SetKwInOut{Input}{input}
    \SetKwInOut{Output}{output}
      \caption{ComputeCosts}\label{alg:supernode_cost}
    
    \Input{{Graph $G' = (\mathcal{V}', \mathcal{E}', \mathbf{X}')$, Node rep. $\tilde{\mathbf{H}}$, Node pairs $\texttt{candidates}$}}
    \Output{Supernode cost map $\texttt{supernode\_costs}$}
    
    \BlankLine
    
    $\texttt{supernode\_costs} \gets Map()$\\
    \For{$(u, v) \in \texttt{candidates}$}{
        $\texttt{\_}, \mathbf{P}, \tilde{\mathbf{H}}', \texttt{\_} \gets \texttt{Merge}(\mathcal{V}', \mathcal{E}', \mathbf{X}', \{(u, v)\})$
        
        $\texttt{supernode\_costs}[(u, v)] \gets \Vert \mathbf{P} \tilde{\mathbf{H}}' - \tilde{\mathbf{H}} \Vert_{1}^{1}$
    }
    \label{alg}
\end{algorithm}

By definition of the GCN and coarse graph convolution operations, the effect of merging $u$ and $v$ can only reach the one-hop neighborhood of the two supernodes.
Let $\mathbf{h}_{i}$ be the representation of node $i$ at the level $l$ of coarsening: $\mathbf{h}_{i} \triangleq \tilde{\mathbf{H}}^{(1)}_{(l)}[i]$.
Additionally, define $\mathbf{h}'_{i}$ to be the representation of node $i$ after merging supernodes $u$ and $v$: $\mathbf{h}'_{i} \triangleq \mathbf{P}_{(u, v)} \tilde{\mathbf{H}}^{(1)}_{(l, \mathbf{P}_{(u, v)})}[i]$.
With this notation, the cost function is equivalently:
{
\begin{align}
    cost(u, v) = \Vert \mathbf{h}_{u} - \mathbf{h}'_{(u, v)} \Vert_{1}^{1} + \Vert \mathbf{h}_{v} - \mathbf{h}'_{(u, v)} \Vert_{1}^{1} + \sum_{i \in \mathcal{N}(\{u, v\})} \Vert \mathbf{h}_{i} - \mathbf{h}'_{i} \Vert_{1}^{1}
\end{align}
}%
where, $\mathbf{h}'_{(u, v)}$ is the representation of the supernode created by the merge.
\algref{alg:supernode_cost} details the cost computation process for \shortname{}.
Additionally, \figref{fig:supernodeCosts} illustrates the cost computation for a small subgraph.

\section{Extended Step 3: Merging Nodes}
\label{app:merging_nodes}

The top-$k$ non-overlapping lowest-cost candidate supernodes are merged simultaneously at every level of coarsening.
For each pair of nodes, $(u, v)$, being merged, the set of edges incident with the resulting supernode is inherited from the nodes $u$ and $v$.
Specifically, let $\textrm{src} = \{w \in \mathcal{V} \, \vert \, (w, u) \in \mathcal{E}_{l} \vee (w, v) \in \mathcal{E}_{l}\}$ and $\textrm{dst} = \{w \in \mathcal{V} \, \vert \, (u, w) \in \mathcal{E}_{l} \vee (v, w) \in \mathcal{E}_{l}\}$, where $\mathcal{E}_{l}$ is the edge set of the graph at the coarsening level $l$.
Then the edges incident with the supernode created by merging $u, v$ is:
$
    \{ (s, w) \, \vert \, w \in \textrm{dst} \} \, \cup \, \{ (w, s) \, \vert \, w \in \textrm{src} \} \nonumber
$.
Furthermore, the edges connecting supernodes in the resulting coarsened graph are weighted by the number of edges connecting nodes in the two incident supernodes.
Recall, if $\mathbf{P}$ is the partitioning matrix representing the assignment of nodes to supernodes, then the coarsened graph's weighted adjacency matrix is $\mathbf{A}' = \mathbf{P}^T \mathbf{A} \mathbf{P}$.
Moreover, the features of the supernodes are a weighted average of the features of the nodes being merged: $\mathbf{X}' = \mathbf{P}^T \mathbf{C}^{-1} \mathbf{X}$.

When two nodes are merged, the merge-graph is also updated. 
As in the coarsened graph, when supernodes $u$ and $v$ are merged to create a new supernode, the new supernode is connected to every neighbor of $u$ and $v$.
This is illustrated in \figref{fig:scalingGNNsViaGraphCoarsening}.
If $\mathbf{A}_{merge, l}$ is the adjacency matrix of the merge-graph at the coarsening level, then after partitioning nodes by $\mathbf{P}_{(l + 1)}$, we have $\mathbf{A}_{merge, l + 1}' = \mathbf{P}_{(l + 1)}^T \mathbf{A}_{merge, l} \mathbf{P}_{(l + 1)}$.
However, the weights of the merge-graph adjacency matrix are irrelevant to the coarsening algorithm.
In addition to updating the structure of the merge-graph after a merge, costs must also be recomputed.
A scalable method for updating costs is described in \appref{app:cost_approximation}.

\section{Caching Node Summation Terms}
\label{app:caching_node_summation_terms}

In this section, we introduce a technique for scaling the exact supernode cost computation.
By definition, the representation of a supernode $i$ obtained via a single layer of coarse GCN convolution is 
\begin{align}
    \mathbf{h}_{i} = \frac{\vert C_{i} \vert}{d_{i} + \vert C_{i} \vert} \mathbf{x}_{i} + \frac{1}{\sqrt{d_{i} + \vert C_{i} \vert}} \sum_{j \in \mathcal{N}(\{i\})} \frac{a_{j i}}{\sqrt{d_{j} + \vert C_{j} \vert}} \mathbf{x}_{j}
\end{align}
where $d_{i}$ is the degree of node $i$, $\mathbf{x}_{i}$ is the attributes of node $i$, and $a_{j i}$ is the adjacency matrix entry for row $j$ column $i$, i.e., the weight of the edge from node $j$ to $i$. 
Define $\mathbf{s}_{i} \triangleq \sum_{j \in \mathcal{N}(\{i\})} \frac{a_{j i}}{\sqrt{d_{j} + \vert C_{j} \vert}} \mathbf{x}_{j}$.
Then, for the nodes being merged, $u$ and $v$, the new representation after the merge is 
{\small
\begin{align}
    \mathbf{h}'_{(u, v)} & = \frac{\vert C_{u} \vert + \vert C_{v} \vert}{d_{i} + \vert C_{u} \vert + \vert C_{v} \vert} \mathbf{x}_{(u, v)} + \frac{(\mathbf{s}_{u} - \frac{a_{v, u}}{\sqrt{d_{v} + \vert C_{v} \vert}} \mathbf{x}_{v} + s_{v} - \frac{a_{u, v}}{\sqrt{d_{u} + \vert C_{u} \vert}}\mathbf{x}_{u})}{\sqrt{d_{(u, v)} + \vert C_{u} \vert + \vert C_{v} \vert}}
\end{align}
}%
where $\mathbf{x}_{(u, v)}$ is the attributes of the supernode created by merging $u$ and $v$.
Observe that if the value of $\mathbf{s}_{i}$ is cached for each node, then the coarse graph convolution output of a supernode created by merging a pair of nodes does not require information from the node neighbors.

Similarly, using the cached value of $\mathbf{s}_{i}$, the new representation after the merge for a node $ i \in \mathcal{N}(\{u, v\}) $ is simplified to 
{\small
\begin{align}
    \mathbf{h}'_{i} =& \frac{\vert C_{i} \vert}{d_{i} + \vert C_{i} \vert} \mathbf{x}_{i} + \frac{1}{\sqrt{d_{i} + \vert C_{i} \vert}} \mathbf{s}_{i} + \frac{a_{u i} + a_{v i}}{\sqrt{(d_{i} + \vert C_{i} \vert)(d_{(u, v)} + \vert C_{u} \vert + \vert C_{v} \vert)}}\mathbf{x}_{(u, v)} \nonumber\\ 
    & - \frac{a_{u i}}{\sqrt{(d_{i} + \vert C_{i} \vert)(d_{u} + \vert C_{u} \vert)}}\mathbf{x}_{u} - \frac{a_{v i}}{\sqrt{(d_{i} + \vert C_{i} \vert)(d_{v} + \vert C_{v} \vert)}}\mathbf{x}_{v} 
\label{eq:neighbor_node_rep_update}
\end{align}
}%
The benefit of caching $\mathbf{s}_{i}$ is that no information from the 2-hop neighbors of the nodes being considered for merging needs to be obtained.

The cached statistics for each node are therefore updated using the following rules:
\begin{align}
    \mathbf{s}_{(u, v)} &= \mathbf{s}_{u} - \frac{a_{v, u}}{\sqrt{d_{v} + \vert C_{v} \vert}}\mathbf{x}_{v} + \mathbf{s}_{v} - \frac{a_{u, v}}{\sqrt{d_{u} + \vert C_{u} \vert}} \mathbf{x}_{u}
\end{align}
Note that the influence and sum statistics of the neighbors of the merged nodes must also be updated as the graph's structure is updated.
{\footnotesize
\begin{align}
    \mathbf{s}'_{i} &= \mathbf{s}_{i} - \frac{a_{u, i}}{\sqrt{d_{u} + \vert C_{u} \vert}} \mathbf{x}_{u} - \frac{a_{v, i}}{\sqrt{d_{v} + \vert C_{v} \vert}}\mathbf{x}_{v} + \frac{a_{u, i} + a_{v, i}}{\sqrt{d_{u} + d_{v} + \vert C_{u} \vert + \vert C_{v} \vert}} \mathbf{x}_{(u, v)}
\end{align}
}%

\section{A Supernode Cost Approximation}
\label{app:cost_approximation}

An approximation yielding significant improvements in graph summarization time is motivated by the upper bound on the merge cost approximation stated in \theoremref{thm:merge_cost_upper_bound}. 
This section provides the proof for this theorem.
\begin{proof}[Proof of \theoremref{thm:merge_cost_upper_bound}]
    Let $\mathbf{h}_{i}$ be the representation of node $i$ at the level $l$ of coarsening: $\mathbf{h}_{i} \triangleq \tilde{\mathbf{H}}^{(1)}_{(l)}[i]$.
    Additionally, define $\mathbf{h}'_{i}$ to be the representation of node $i$ after merging supernodes $u$ and $v$: $\mathbf{h}'_{i} \triangleq \mathbf{P}_{(u, v)} \tilde{\mathbf{H}}^{(1)}_{(l, \mathbf{P}_{(u, v)})}[i]$.
    Starting from the definition of the supernode cost provided in \eqref{eq:merge_cost}, we have:
    {\small
    \begin{align}
        cost(u, v) = \Vert \mathbf{h}_{u} - \mathbf{h}'_{(u, v)} \Vert_{1}^{1} + \Vert \mathbf{h}_{v} - \mathbf{h}'_{(u, v)} \Vert_{1}^{1} + \sum_{i \in \mathcal{N}(\{u, v\})} \Vert \mathbf{h}_{i} - \mathbf{h}'_{i} \Vert_{1}^{1}
    \end{align}
    }%
    By the definitions of $\mathbf{h}_{i}$ and $\mathbf{h}'_{i}$ provided in the previous section, the $ \Vert \mathbf{h}_{i} - \mathbf{h}'_{i} \Vert_{1}^{1} $ terms in the summation can be expanded and simplified
    {\footnotesize
    \begin{align}
        & \Vert \mathbf{h}_{i} - \mathbf{h}'_{i} \Vert_{1}^{1} = \Vert \frac{\vert C_{i} \vert}{d_{i} + \vert C_{i} \vert} \mathbf{x}_{i} + \frac{1}{\sqrt{d_{i} + \vert C_{i} \vert}} \sum_{j \in \mathcal{N}(\{i\})} \frac{a_{j i}}{\sqrt{d_{j} + \vert C_{j} \vert}} \mathbf{x}_{j} \nonumber \\
        & \quad - \frac{\vert C_{i} \vert}{d_{i} - \vert C_{i} \vert} \mathbf{x}_{i} - \frac{1}{\sqrt{d_{i} + \vert C_{i} \vert}} \sum_{j \in \mathcal{N}(\{i\}) \setminus \{ u, v \}} \frac{a_{j i}}{\sqrt{d_{j} + \vert C_{j} \vert}} \mathbf{x}_{j} \nonumber\\
        & \quad - \frac{a_{u, i} + a_{v, i}}{\sqrt{(d_{i} + \vert C_{i} \vert)(d_{(u,v)} + \vert C_{u} \vert + \vert C_{v} \vert)}}\mathbf{x}_{(u, v)} \Vert_{1}^{1} \nonumber \\
        & = \Vert \frac{a_{u, i}}{\sqrt{(d_{i} + \vert C_{i} \vert)(d_{u} + \vert C_{u} \vert)}}\mathbf{x}_{u} 
        + \frac{a_{v, i}}{\sqrt{(d_{i} + \vert C_{i} \vert)(d_{v} + \vert C_{v} \vert)}}\mathbf{x}_{v} \nonumber\\ 
        & \quad - \frac{a_{u, i} + a_{v, i}}{\sqrt{(d_{i} + \vert C_{i} \vert)(d_{(u,v)} + \vert C_{u} \vert + \vert C_{v} \vert)}}\mathbf{x}_{(u, v)} \Vert_{1}^{1}
        \label{eq:representation_change}
    \end{align}
    }%
    Then, using the sub-additivity and absolute homogeneity properties of norms we realize the following inequality.
    {\small
    \begin{align}
        &= \Vert \frac{a_{u, i}}{\sqrt{(d_{i} + \vert C_{i} \vert)(d_{u} + \vert C_{u} \vert)}}\mathbf{x}_{u} 
        + \frac{a_{v, i}}{\sqrt{(d_{i} + \vert C_{i} \vert)(d_{v} + \vert C_{v} \vert)}}\mathbf{x}_{v} \nonumber\\ 
        & \quad - \frac{a_{u, i} + a_{v, i}}{\sqrt{(d_{i} + \vert C_{i} \vert)(d_{(u,v)} + \vert C_{u} \vert + \vert C_{v} \vert)}}\mathbf{x}_{(u, v)} \Vert_{1}^{1} \nonumber\\
        & \leq \frac{a_{u, i}}{\sqrt{(d_{i} + \vert C_{i} \vert)}} \Vert \frac{1}{\sqrt{(d_{(u,v)} + \vert C_{u} \vert + \vert C_{v} \vert)}}\mathbf{x}_{(u, v)} - \frac{1}{\sqrt{(d_{u} + \vert C_{u} \vert)}}\mathbf{x}_{u} \Vert_1^1 \nonumber \\
        & \quad + \frac{a_{v, i}}{\sqrt{(d_{i} + \vert C_{i} \vert)}} \Vert \frac{1}{\sqrt{(d_{(u,v)} + \vert C_{u} \vert + \vert C_{v} \vert)}}\mathbf{x}_{(u, v)} - \frac{1}{\sqrt{(d_{v} + \vert C_{v} \vert)}}\mathbf{x}_{v} \Vert_1^1
    \end{align}
    \label{eq:instance_of_cost_bound}
    }%
    Observe if one of $a_{v, i}$ or $a_{u, i}$ is $0$, the inequality is satisfied with equality.
    Plugging this result into the definition of supernode costs yields our upper bound.
    {\footnotesize
    \begin{align}
        cost&(u, v) \leq \Vert \mathbf{h}_{u} - \mathbf{h}'_{(u, v)} \Vert_{1}^{1} + \Vert \mathbf{h}_{v} - \mathbf{h}'_{(u, v)} \Vert_{1}^{1} \nonumber\\ 
        &+ \Vert \frac{\mathbf{x}_{(u, v)}}{\sqrt{(d_{(u,v)} + \vert C_{u} \vert + \vert C_{v} \vert)}} - \frac{\mathbf{x}_{u}}{\sqrt{(d_{u} + \vert C_{u} \vert)}} \Vert_1^1 \sum_{i \in \mathcal{N}(\{u\})} \frac{a_{u, i}}{\sqrt{(d_{i} + \vert C_{i} \vert)}} \nonumber\\
        &+ \Vert \frac{\mathbf{x}_{(u, v)}}{\sqrt{(d_{(u,v)} + \vert C_{u} \vert + \vert C_{v} \vert)}} - \frac{\mathbf{x}_{v}}{\sqrt{(d_{v} + \vert C_{v} \vert)}} \Vert_1^1 \sum_{i \in \mathcal{N}(\{v\})} \frac{a_{v, i}}{\sqrt{(d_{i} + \vert C_{i} \vert)}}
    \end{align}
    }%
    In the case $\mathcal{N}(u) \cap \mathcal{N}(v) = \emptyset$, every instance of the bound in \eqref{eq:instance_of_cost_bound} is satisfied with equality, and consequentially the bound above is satisfied with equality.
\end{proof}

As a result, the following term, referred to as the node's \emph{influence}, is cached for all nodes $v \in \mathcal{V}$: $\textit{infl}_{v} \ \triangleq \sum_{i \in \mathcal{N}(\{v\})} \frac{a_{v i}}{\sqrt{(d_{i} + \vert C_{i} \vert)}}.$
This allows the cost of merging two nodes to be a function of properties local to the two nodes being considered, making the cost computation fast and highly scalable. 

In addition to updating the structure of the merge-graph after a merge, the costs and the cached influence scores for each node must be updated.
The influence of the supernode created by merging two nodes $u$ and $v$ is:
{
\begin{align}
    \textit{infl}_{l, (u, v)} &= \left ( \textit{infl}_{l, u} - \frac{a_{u v}}{\sqrt{d_{v} + \vert C_{v} \vert}} \right ) + \left ( \textit{infl}_{v} - \frac{a_{v, u}}{\sqrt{d_{u} + \vert C_{u} \vert}} \right ).
\end{align}
}%
Furthermore, the influence score of each neighbor, $i$, of two merged nodes $u$ and $v$ is:
{
\begin{align}
    \textit{infl}_{l, i} &= \textit{infl}_{l, i} - \frac{a_{i u}}{\sqrt{d_{u} + \vert C_{u} \vert}} - \frac{a_{i v}}{\sqrt{d_{v} + \vert C_{v} \vert}} + \frac{a_{u i} + a_{v i}}{\sqrt{d_{u} + d_{v} + \vert C_{u} \vert + \vert C_{v} \vert}}.
\end{align}
}%

\section{Extended Evaluation}
\label{app:extended_evaluation}

\subsection{Graph Summarization Time}

\input{figures/fullGraphSummarizationTimes}

We compare the graph summarization time of \shortname{} and A-\shortname{} to baselines at varying coarsening ratios for each dataset and task.
The average time across 5 rounds of summarization for all datasets are shown in \figref{fig:fullgraphSummarizationTimes}.
A-\shortname{} is consistently faster than all other baseline graph summarizers on the larger OGB datasets.
This is partially explained by the fact that much of the A-\shortname{} algorithm is highly parallelizable due to our cost approximation.

\par{\textbf{Baselines}.}
The RS method randomly samples nodes from the original graph and uses the induced subgraph to train the GNN.
Herding and KCenter first fit node embeddings for the NC task and then group the nodes by labels. 
The Herding and KCenter methods then select nodes from each group to create a subgraph.
Herding and KCenter require a class label to group nodes are therefore only used in NC settings.
Furthermore, the implementation of Herding and KCenter follows that of \citep{jin:kdd22}.
The original implementation was extended to reach coarsening ratios exceeding the training labeling rate of the dataset by treating the the unlabeled nodes as a distinct class in such settings.
GCond \citep{jin:iclr22} and DosCond \citep{jin:kdd22} train a GNN on the original graph and fit synthetic graph features and connections so the gradient with respect to the GNN weights computed with both graphs are similar.
We use the implementation of GCond and DosCond provided in \citep{jin:kdd22} for NC tasks and extend their method for LP using an appropriate link prediction training loss.
Furthermore, to support coarsening ratios exceeding the training labeling rate of the dataset, synthetic node features are initialized using a sampling procedure with replacement.
Finally, the VN approach is a coarsening algorithm proposed by \citenoun{loukas:jmlr19} that recursively merges neighborhoods of nodes into supernodes.
\citenoun{huang:kdd21} found VN resulted in the best overall prediction performance of the coarsening algorithms proposed by \citenoun{loukas:jmlr19}.
We use the implementation of VN provided by \citenoun{huang:kdd21}.

\begin{table*}[h]
    \centering
    \begin{tabular}{c||c|c|c|c|c}
    \toprule
        \textbf{Dataset} & \textbf{Num Layers} & \textbf{Hidden Dim} & \textbf{Learning Rate} & \textbf{Dropout} & \textbf{Weight Decay} \\
    \midrule
    \midrule
        Citeseer LP & 2 & 256 & 1.0e-2 & 0.5 & 0.0 \\
        \hline
        Cora LP & 2 & 256 & 1.0e-2 & 0.5 & 0.0 \\
        \hline
        OGBLCol & 3 & 256 & 1.0e-3 & 0.0 & 0.0 \\
        \hline
        OGBLCit2 & 3 & 256 & 5.0e-4 & 0.0 & 0.0 \\
    \midrule
    \midrule
        Citeseer NC & 2 & 256 & 1.0e-2 & 0.5 & 5.0e-4 \\
        \hline
        Cora NC & 2 & 256 & 1.0e-2 & 0.5 & 5.0e-4 \\
        \hline
        OGBNArxiv & 3 & 256 & 1.0e-2 & 0.5 & 0.0 \\
        \hline
        OGBNProd & 3 & 256 & 1.0e-2 & 0.5 & 0.0 \\
    \bottomrule
    \end{tabular}
    \caption{Table of GCN network and training hyperparameters.}
    \label{tab:gcn_params}
\end{table*}

\begin{table*}[t]
    \centering
    \begin{tabular}{c||c|c|c|c|c}
    \toprule
        \textbf{Dataset} & \textbf{Merge Batch Size} & \textbf{SGC}-K & \textbf{PCA Dim} & Top-$k_{nn}$ & $d_{nn}$ \\
    \midrule
    \midrule
        Citeseer LP & 1 & [2, 3, 4*] & [ 5, 10*, 15 ] & [ 1*, 3 ] & [ 0.01*, 0.1, 0.5 ] \\
        \hline
        Cora LP & 10 & [2, 3, 4*] & [ 5, 10, 15* ] & [ 1*, 3 ] & [ 0.01*, 0.1, 0.5 ]\\
        \hline
        OGBLCol & 10,000 & [2*, 3, 4] & [ 10, 20* ] & [ 1*, 3 ] & [ 0.001*, 0.01, 0.1 ]\\
        \hline
        OGBLCit2 & 10,000 & [2*, 3] & [ 10*, 20 ] & [ 1*, 3 ] & [ 0.001*, 0.01 ] \\
    \midrule
    \midrule
        Citeseer NC & 1 & [2, 3*, 4] & [ 5*, 10, 15 ] & [ 1, 3* ] & [ 0.01, 0.1*, 0.5 ] \\
        \hline
        Cora NC & 10 & [2, 3*, 4] & [ 5, 10, 15* ] & [ 1*, 3 ] & [ 0.01*, 0.1, 0.5 ] \\
        \hline
        OGBNArxiv & 10,000 & [2, 3*, 4] & [ 10, 20* ] & [ 1*, 3 ] & [ 0.001, 0.01*, 0.1 ] \\
        \hline
        OGBNProd & 100,000 & [2*, 3] & [ 10*, 20 ] & [ 1*, 3 ] & [ 0.001*, 0.01 ] \\
    \bottomrule
    \end{tabular}
    \caption{Table of \shortname{} hyperparameters.}
    \label{tab:conv_match_param_search}
\end{table*}

\par{\textbf{Model Architectures and Hyperparameters}.} 
The GCN architectures and training parameters for Citeseer and Cora are from \cite{jin:kdd22} and the GCN architectures and training parameters for OGB datasets are from \cite{hu:nips20}.
Every GCN is trained using the ADAM optimizer implementation from PyTorch.
The \tabref{tab:gcn_params} summarizes the parameters used in the experiments.

The hyperparameters for the \shortname and A-\shortname algorithms were tuned for each dataset using validation data at a summarization rate $r = 1\%$ and at a selected batch size.
The \tabref{tab:conv_match_param_search} summarizes the final parameters used in the experiments.
Hyperparameter settings for VN and the two coreset baselines (Herding and KCenter) are taken from \cite{huang:kdd21} and \cite{jin:kdd22}, respectively.
Hyperparameter settings for the GCond and DosCond methods are taken from \cite{jin:kdd22} on the datasets they examined, otherwise, they are found via a hyperparameter search.

\noindent \textbf{Hardware}.
All experiments were run on an AWS p3.16xlarge EC2 instance with 8 16GB NVIDIA Tesla V100 GPUs.

%% file: figures/fullGraphSummarizationTimes.tex
\begin{figure}
\begin{subfigure}{.25\textwidth}
  \centering
  \includegraphics[width=.99\linewidth]{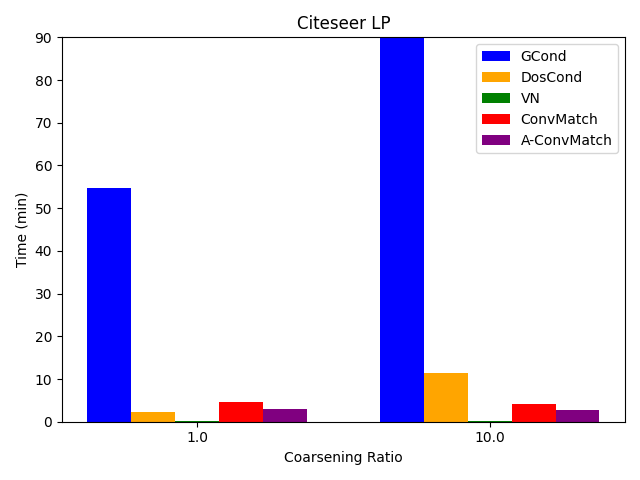}
  \label{fig:fullciteseerLPGraphSummarizationTime}
\end{subfigure}%
\begin{subfigure}{.25\textwidth}
  \centering
  \includegraphics[width=.99\linewidth]{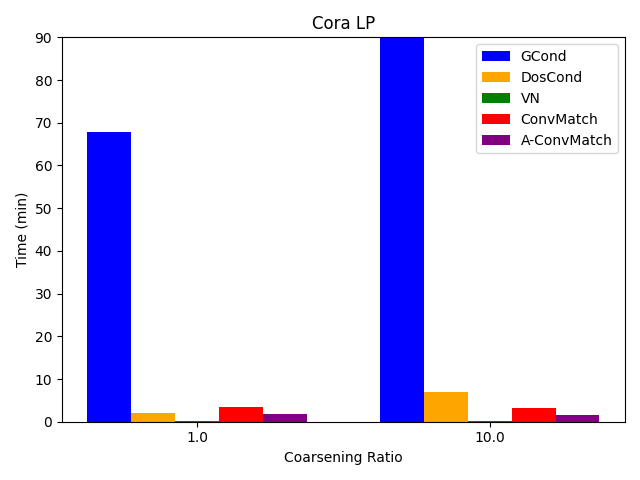}
  \label{fig:fullcoraLPGraphSummarizationTime}
\end{subfigure}%
\begin{subfigure}{.25\textwidth}
  \centering
  \includegraphics[width=.99\linewidth]{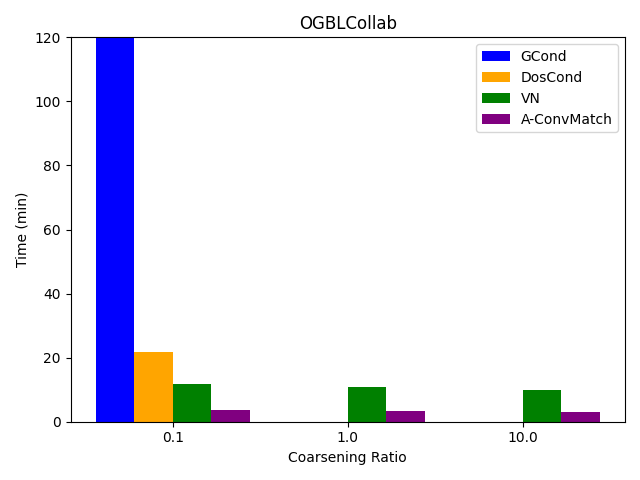}
  \label{fig:fullOGBLCollabLPGraphSummarizationTime}
\end{subfigure}%
\begin{subfigure}{.25\textwidth}
  \centering
  \includegraphics[width=.99\linewidth]{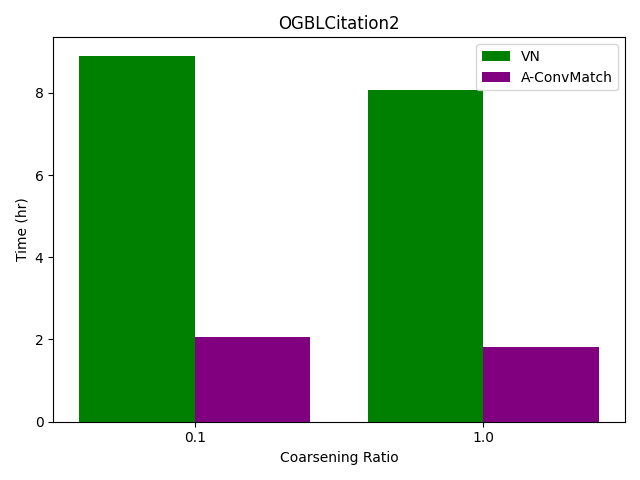}
  \label{fig:fullOGBLCitation2LPGraphSummarizationTime}
\end{subfigure}
\begin{subfigure}{.25\textwidth}
  \centering
  \includegraphics[width=.99\linewidth]{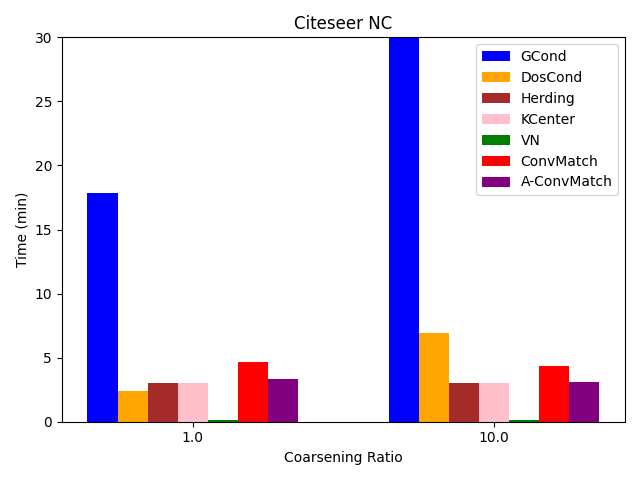}
  \label{fig:fullciteseerNCGraphSummarizationTime}
\end{subfigure}%
\begin{subfigure}{.25\textwidth}
  \centering
  \includegraphics[width=.99\linewidth]{figures/CoraNCGraphSummarizationTime.png}
  \label{fig:fullcoraNCGraphSummarizationTime}
\end{subfigure}%
\begin{subfigure}{.25\textwidth}
  \centering
  \includegraphics[width=.99\linewidth]{figures/OGBNArxivGraphSummarizationTime.png}
  \label{fig:fullOGBNArxivGraphSummarizationTime}
\end{subfigure}%
\begin{subfigure}{.25\textwidth}
  \centering
  \includegraphics[width=.99\linewidth]{figures/OGBNProductsGraphSummarizationTime.png}
  \label{fig:fullOGBNProductsGraphSummarizationTime}
\end{subfigure}
\caption{Plots of graph summarization times at multiple coarsening ratios for all datasets and tasks.
\shortname{} and A-\shortname{} are fast summarization algorithms when compared to baselines. 
}
\label{fig:fullgraphSummarizationTimes}
\end{figure}

%% file: main.bbl
\begin{thebibliography}{45}
\providecommand{\natexlab}[1]{#1}
\providecommand{\url}[1]{\texttt{#1}}
\expandafter\ifx\csname urlstyle\endcsname\relax
  \providecommand{\doi}[1]{doi: #1}\else
  \providecommand{\doi}{doi: \begingroup \urlstyle{rm}\Url}\fi

\bibitem[Hu et~al.(2020{\natexlab{a}})Hu, Fey, Zitnik, Dong, Ren, Liu, Catasta,
  and Leskovec]{hu2020ogb}
Weihua Hu, Matthias Fey, Marinka Zitnik, Yuxiao Dong, Hongyu Ren, Bowen Liu,
  Michele Catasta, and Jure Leskovec.
\newblock Open graph benchmark: Datasets for machine learning on graphs.
\newblock \emph{Advances in neural information processing systems},
  33:\penalty0 22118--22133, 2020{\natexlab{a}}.

\bibitem[Wang et~al.(2019)Wang, Zheng, Ye, Gan, Li, Song, Zhou, Ma, Yu, Gai,
  Xiao, He, Karypis, Li, and Zhang]{wang2019dgl}
Minjie Wang, Da~Zheng, Zihao Ye, Quan Gan, Mufei Li, Xiang Song, Jinjing Zhou,
  Chao Ma, Lingfan Yu, Yu~Gai, Tianjun Xiao, Tong He, George Karypis, Jinyang
  Li, and Zheng Zhang.
\newblock Deep graph library: A graph-centric, highly-performant package for
  graph neural networks.
\newblock \emph{arXiv preprint arXiv:1909.01315}, 2019.

\bibitem[Fey and Lenssen(2019)]{Fey2019PyG}
Matthias Fey and Jan~E. Lenssen.
\newblock Fast graph representation learning with {PyTorch Geometric}.
\newblock In \emph{ICLR Workshop on Representation Learning on Graphs and
  Manifolds}, 2019.

\bibitem[Zheng et~al.(2021)Zheng, Song, Yang, LaSalle, and
  Karypis]{zheng2021distributed}
Da~Zheng, Xiang Song, Chengru Yang, Dominique LaSalle, and George Karypis.
\newblock Distributed hybrid cpu and gpu training for graph neural networks on
  billion-scale graphs.
\newblock \emph{arXiv preprint arXiv:2112.15345}, 2021.

\bibitem[Zheng et~al.(2020)Zheng, Ma, Wang, Zhou, Su, Song, Gan, Zhang, and
  Karypis]{zheng2020distdgl}
Da~Zheng, Chao Ma, Minjie Wang, Jinjing Zhou, Qidong Su, Xiang Song, Quan Gan,
  Zheng Zhang, and George Karypis.
\newblock Distdgl: distributed graph neural network training for billion-scale
  graphs.
\newblock In \emph{2020 IEEE/ACM 10th Workshop on Irregular Applications:
  Architectures and Algorithms (IA3)}, pages 36--44. IEEE, 2020.

\bibitem[Jiang and Rumi(2021)]{jiang2021communication}
Peng Jiang and Masuma~Akter Rumi.
\newblock Communication-efficient sampling for distributed training of graph
  convolutional networks.
\newblock \emph{arXiv preprint arXiv:2101.07706}, 2021.

\bibitem[Ramezani et~al.(2021)Ramezani, Cong, Mahdavi, Kandemir, and
  Sivasubramaniam]{ramezani2021learn}
Morteza Ramezani, Weilin Cong, Mehrdad Mahdavi, Mahmut Kandemir, and Anand
  Sivasubramaniam.
\newblock Learn locally, correct globally: A distributed algorithm for training
  graph neural networks.
\newblock In \emph{International Conference on Learning Representations}, 2021.

\bibitem[Zhu et~al.(2023)Zhu, Reganti, Huang, Dickens, Rao, Subbian, and
  Koutra]{zhu2023simplifying}
Jiong Zhu, Aishwarya~Naresh Reganti, Edward~W Huang, Charles~Andrew Dickens,
  Nikhil Rao, Karthik Subbian, and Danai Koutra.
\newblock Simplifying distributed neural network training on massive graphs:
  Randomized partitions improve model aggregation.
\newblock In \emph{ICML Workshop on Localized Learning (LLW)}, 2023.

\bibitem[Gandhi and Iyer(2021)]{gandhi2021p3}
Swapnil Gandhi and Anand~Padmanabha Iyer.
\newblock P3: Distributed deep graph learning at scale.
\newblock In \emph{15th $\{$USENIX$\}$ Symposium on Operating Systems Design
  and Implementation ($\{$OSDI$\}$ 21)}, pages 551--568, 2021.

\bibitem[{W. Tsang} et~al.(2005){W. Tsang}, {T. Kwok}, and
  Cheung]{tsang:jmlr05}
Ivor {W. Tsang}, James {T. Kwok}, and Pak-Min Cheung.
\newblock Core vector machines: Fast svm training on very large data sets.
\newblock \emph{Journal of Machine Learning Research}, 6:\penalty0 363--392,
  2005.

\bibitem[Har-Peled and Kushal(2005)]{har-peled:acg05}
Sariel Har-Peled and Akash Kushal.
\newblock Smaller coresets for k-mediam and k-means clustering.
\newblock In \emph{ACM Annual Symposium on Computational Geometry (ACG)}, 2005.

\bibitem[Welling(2009)]{welling:icml09}
Max Welling.
\newblock Herding dynamical weights to learn.
\newblock In \emph{International Conference on Machine Learning (ICML)}, 2009.

\bibitem[Sener and Savarses(2018)]{sener:iclr18}
Ozan Sener and Silvio Savarses.
\newblock Active learning for convolutional neural networks: A core-set
  approach.
\newblock In \emph{International Conference on Learning Representations
  (ICLR)}, 2018.

\bibitem[Mirzasoleiman et~al.(2020)Mirzasoleiman, Bilmes, and
  Leskovec]{mirzasoleiman:icml20}
Baharan Mirzasoleiman, Jeff Bilmes, and Jure Leskovec.
\newblock Coreset for data-efficient training of machine learning models.
\newblock In \emph{International Conference on Machine Learning (ICML)}, 2020.

\bibitem[Huang et~al.(2021)Huang, Zhang, Xi, Liu, and Zhou]{huang:kdd21}
Zengfeng Huang, Shengzhong Zhang, Chong Xi, Tang Liu, and Min Zhou.
\newblock Scaling up graph neural networks via graph coarsening.
\newblock In \emph{ACM SIGKDD International Conference on Knowledge Discovery
  \& Data Mining (KDD)}, 2021.

\bibitem[Zhao et~al.(2021)Zhao, {Reddy Mopuri}, and Bilen]{zhao:iclr21}
Bo~Zhao, Konda {Reddy Mopuri}, and Hakan Bilen.
\newblock Dataset condensation with gradient matching.
\newblock In \emph{International Conference on Learning Representations
  (ICLR)}, 2021.

\bibitem[Jin et~al.(2022{\natexlab{a}})Jin, Tang, Jiang, Li, Zhang, Tang, and
  Yin]{jin:kdd22}
Wei Jin, Xianfeng Tang, Haoming Jiang, Zheng Li, Danqing Zhang, Jiliang Tang,
  and Bing Yin.
\newblock Condensing graphs via one-step gradient matching.
\newblock In \emph{ACM SIGKDD International Conference on Knowledge Discovery
  \& Data Mining (KDD)}, 2022{\natexlab{a}}.

\bibitem[Jin et~al.(2022{\natexlab{b}})Jin, Zhao, Zhang, Liu, Tang, and
  Shah]{jin:iclr22}
Wei Jin, Lingxiao Zhao, Schichang Zhang, Yozen Liu, Jiliang Tang, and Neil
  Shah.
\newblock Graph condensation for graph neural networks.
\newblock In \emph{International Conference on Learning Representations
  (ICLR)}, 2022{\natexlab{b}}.

\bibitem[Loukas(2019)]{loukas:jmlr19}
Andreas Loukas.
\newblock Graph reduction with spectral and cut guarantees.
\newblock \emph{Journal of Machine Learning Research}, 20\penalty0
  (116):\penalty0 1--42, 2019.

\bibitem[Wang et~al.(2018)Wang, Zhu, Torralba, and {A. Efros}]{wang:arxiv18}
Tongzhou Wang, {Jun-Yan} Zhu, Antonio Torralba, and Alexei {A. Efros}.
\newblock Dataset distillation.
\newblock In \emph{Arxiv preprint}, 2018.

\bibitem[Bohdal et~al.(2020)Bohdal, Yang, and Hospedales]{bohdal:arxiv20}
Ondrej Bohdal, Yongxin Yang, and Timothy Hospedales.
\newblock Flexible dataset distillation: Learn labels instead of images.
\newblock In \emph{Arxiv preprint}, 2020.

\bibitem[Sucholutsky and Schonlau(2020)]{sucholutsky:arxiv20}
Ilia Sucholutsky and Matthias Schonlau.
\newblock Soft-label dataset distillation and text dataset distillation.
\newblock In \emph{Arxiv preprint}, 2020.

\bibitem[Nguyen et~al.(2021)Nguyen, Chen, and Lee]{nguyen:iclr21}
Timothy Nguyen, Zhourong Chen, and Jaehoon Lee.
\newblock Dataset meta-learning from kernel ridge-regression.
\newblock In \emph{International Conference on Learning Representations
  (ICLR)}, 2021.

\bibitem[Zhao and Bilen(2021)]{zhao:icml21}
Bo~Zhao and Hakan Bilen.
\newblock Dataset condensation with differentiable siamese augmentation.
\newblock In \emph{International Conference on Machine Learning (ICML)}, 2021.

\bibitem[Liu et~al.(2022)Liu, Li, Chen, and Song]{liu:arxiv22}
Mengyang Liu, Shanchuan Li, Xinshi Chen, and Le~Song.
\newblock Graph condensation via receptive field distribution matching.
\newblock In \emph{Arxiv preprint}, 2022.

\bibitem[Liu et~al.(2019)Liu, Safavi, Dighe, and Koutra]{liu:compserv19}
Yike Liu, Tara Safavi, Abhilash Dighe, and Danai Koutra.
\newblock Graph summarization methods and applications: A survey.
\newblock \emph{ACM Computing Surveys}, 51\penalty0 (62), 2019.

\bibitem[Purohit et~al.(2014)Purohit, Prakash, Kang, Zhang, and
  Subrahmanian]{purohit:kdd14}
Manish Purohit, {B. Aditya} Prakash, Chanhyun Kang, Yao Zhang, and {V. S.}
  Subrahmanian.
\newblock Fast influence-based coarsening for large networks.
\newblock In \emph{ACM SIGKDD International Conference on Knowledge Discovery
  \& Data Mining (KDD)}, 2014.

\bibitem[Harel and Koren(2002)]{harel:jgaa02}
David Harel and Yehuda Koren.
\newblock A fast multi-scale method for drawing large graphs.
\newblock \emph{Journal of Graph Algorithms and Applications}, 6\penalty0
  (3):\penalty0 179--202, 2002.

\bibitem[Walshaw(2006)]{walshaw:jgaa06}
Chris Walshaw.
\newblock A multilevel algorithm for force-directed graph drawing.
\newblock \emph{Journal of Graph Algorithms and Applications}, 7\penalty0
  (3):\penalty0 253--285, 2006.

\bibitem[Dunne and Schneiderman(2013)]{shneiderman:sigchi13}
Cody Dunne and Ben Schneiderman.
\newblock Motif simplification: Improving network visualization readability
  with fan, connector, and clique glyphs.
\newblock In \emph{ACM SIGCHI Conference on Human Factors in Computing Systems
  (CHI)}, 2013.

\bibitem[Moitra(2009)]{moitra:sfcs09}
Ankur Moitra.
\newblock Approximation algorithms for multicommodity-type problems with
  guarantees independent of the graph size.
\newblock In \emph{IEEE Symposium on Foundation of Computer Science}, 2009.

\bibitem[Englert et~al.(2014)Englert, Gupta, Krauthgamer, Racke,
  {Talgam-Cohen}, and Talwar]{englert:siamcomp14}
Matthias Englert, Anupam Gupta, Robert Krauthgamer, Harald Racke, Inbal
  {Talgam-Cohen}, and Kunal Talwar.
\newblock Vertex sparsifiers: New results from old techniques.
\newblock \emph{{SIAM} Journal on Computing}, 43\penalty0 (4):\penalty0
  1239--1262, 2014.

\bibitem[Liang et~al.(2021)Liang, Gurukar, and Parthasarathy]{liang:icwsm21}
Jiongqian Liang, Saket Gurukar, and Srinivasan Parthasarathy.
\newblock Mile: A multi-level framework for scalable graph embedding.
\newblock In \emph{International AAAI Conference on Web and Social Media
  (ICWSM)}, 2021.

\bibitem[Akbas and Aktas(2019)]{akbas:ieeebd19}
Esra Akbas and {Mehmet Emin} Aktas.
\newblock Network embedding: On compression and learning.
\newblock In \emph{IEEE Transactions on Big Data}, 2019.

\bibitem[Fahrbach et~al.(2020)Fahrbach, Goranci, Peng, Sachdeva, and
  Wang]{fahrbach:icml20}
Matthew Fahrbach, Gramoz Goranci, Richard Peng, Sushant Sachdeva, and Chi Wang.
\newblock Faster graph embeddings via coarsening.
\newblock In \emph{International Conference on Machine Learning (ICML)}, 2020.

\bibitem[Zhang et~al.(2020)Zhang, Yang, Liu, Sun, Fang, Zhang, and
  Lin]{zhang:ieeekde20}
Zhengyan Zhang, Cheng Yang, Zhiyuan Liu, Maosong Sun, Zhichong Fang, Bo~Zhang,
  and Leyu Lin.
\newblock Cosine: Compressive network embedding on large-scale information
  networks.
\newblock In \emph{IEEE Transactions on Knowledge and Data Engineering}, 2020.

\bibitem[Deng et~al.(2020)Deng, Zhao, Wang, Zhang, and Feng]{deng:iclr20}
Chenhui Deng, Zhiqiang Zhao, Yongyu Wang, Zhiru Zhang, and Zhuo Feng.
\newblock Graphzoom: A multi-level spectral approach for accurate and scalable
  graph embedding.
\newblock In \emph{International Conference on Learning Representations
  (ICLR)}, 2020.

\bibitem[Kumar et~al.(2023)Kumar, Sharma, Saxena, and Kumar]{kumar:icml23}
Manoj Kumar, Anurag Sharma, Shashwat Saxena, and Sandeep Kumar.
\newblock Featured graph coarsening with similarity guarantees.
\newblock In \emph{International Conference on Machine Learning (ICML)}, 2023.

\bibitem[Bruna et~al.(2014)Bruna, Zaremba, Szlam, and Lecun]{bruna:iclr14}
Joan Bruna, Wojciech Zaremba, Arthur Szlam, and Yann Lecun.
\newblock Spectral networks and locally connected networks on graphs.
\newblock In \emph{International Conference on Learning Representations
  (ICLR)}, 2014.

\bibitem[Defferrard et~al.(2016)Defferrard, Bresson, and
  Vandergheynst]{defferrard:nips16}
Michaël Defferrard, Xavier Bresson, and Pierre Vandergheynst.
\newblock Convolutional neural networks on graphs with fast localized spectral
  filtering.
\newblock In \emph{Advances in Neural Information Processing Systems (NIPS)},
  2016.

\bibitem[Kipf and Welling(2017)]{kipf:iclr17}
Thomas~N. Kipf and Max Welling.
\newblock Semi-supervised classification with graph convolutional networks.
\newblock In \emph{International Conference on Learning Representations
  (ICLR)}, 2017.

\bibitem[Levie et~al.(2017)Levie, Monti, Bresson, and Bronstein]{levie:sp18}
Ron Levie, Federico Monti, Xavier Bresson, and Michael~M. Bronstein.
\newblock Cayleynets: Graph convolutional neural networks with complex rational
  spectral filters.
\newblock \emph{IEEE Transactions on Signal Processing}, 67\penalty0
  (1):\penalty0 97--109, 2017.

\bibitem[Wu et~al.(2019)Wu, Zhang, {Holanda de Souza Jr.}, Fifty, Yu, and
  Wienberger]{wu:icml19}
Felix Wu, Tianyi Zhang, Amauri {Holanda de Souza Jr.}, Christopher Fifty, Tao
  Yu, and Kilian~Q. Wienberger.
\newblock Simplifying graph convolutional networks.
\newblock In \emph{International Conference on Machine Learning (ICML)}, 2019.

\bibitem[Sen et~al.(2008)Sen, Namata, Bilgic, Getoor, Galligher, and
  {Eliassi-Rad}]{sen:ai08}
Prithviraj Sen, Galileo Namata, Mustafa Bilgic, Lise Getoor, Brian Galligher,
  and Tina {Eliassi-Rad}.
\newblock Collective classification in network data.
\newblock \emph{AI Magazine}, 29\penalty0 (3), 2008.

\bibitem[Hu et~al.(2020{\natexlab{b}})Hu, Fey, Zitnk, Dong, Ren, Liu, Catasta,
  and Leskovec]{hu:nips20}
Weihua Hu, Matthias Fey, Marinka Zitnk, Yuxiao Dong, Hongyu Ren, Bowen Liu,
  Michele Catasta, and Jure Leskovec.
\newblock Open graph benchmark: Datasets for machine learning on graphs.
\newblock In \emph{Conference on Neural Information Processing Systems
  (NeurIPS)}, 2020{\natexlab{b}}.

\end{thebibliography}
